\documentclass[a4paper,fleqn]{cas-dc}

\usepackage[authoryear]{natbib}

\usepackage{multirow}%
\usepackage{amsmath,amssymb,amsfonts}
\usepackage{mathrsfs}%
\usepackage{listings}%
\usepackage{graphicx}
\usepackage{pifont}
\usepackage{booktabs}
\usepackage{algorithm,algorithmic}
\usepackage{pifont}
\usepackage{tabularx}
\usepackage{hyperref}

\definecolor{mygray}{RGB}{239, 239, 239}

\def\tsc#1{\csdef{#1}{\textsc{\lowercase{#1}}\xspace}}
\tsc{WGM}
\tsc{QE}

\begin{document}
\let\WriteBookmarks\relax
\def\floatpagepagefraction{1}
\def\textpagefraction{.001}

\shorttitle{}    

\shortauthors{}  

\title [mode = title]{UniCAD: Efficient and Extendable Architecture for Multi-Task Computer-Aided Diagnosis System}  



%

\author[1]{Yitao Zhu}


\fnmark[1]




\affiliation[1]{organization={School of Biomedical Engineering \& State Key Laboratory of Advanced Medical Materials and Devices, ShanghaiTech University},
            city={Shanghai},
            postcode={201210}, 
            state={Shanghai},
            country={China}}
\affiliation[2]{organization={School of Biomedical Engineering, Shanghai Jiao Tong University},
            city={Shanghai},
            postcode={200030}, 
            state={Shanghai},
            country={China}}
\affiliation[3]{organization={Shanghai Clinical Research and Trial Center},
            city={Shanghai},
            postcode={201210}, 
            state={Shanghai},
            country={China}}
\affiliation[4]{organization={Shanghai United Imaging Intelligence Co. Ltd.},
            city={Shanghai},
            postcode={200230}, 
            state={Shanghai},
            country={China}}

\author[1]{Yuan Yin}

\fnmark[1]

\author[2]{Zhenrong Shen}

\author[1]{Zihao Zhao}

\author[1]{Haiyu Song}

\author[2]{Sheng Wang}

\author[1,3,4]{Dinggang Shen}

\author[1,3]{Qian Wang}[orcid=0000-0002-3490-3836]
\cormark[1]
\ead{qianwang@shanghaitech.edu.cn}




\cortext[1]{Corresponding author}

\fntext[1]{These authors contributed equally to this work.}


\begin{abstract}
The growing complexity and scale of visual model pre-training have made developing and deploying multi-task computer-aided diagnosis (CAD) systems increasingly challenging and resource-intensive. 
Furthermore, the medical imaging community lacks an open-source CAD platform to enable the rapid creation of efficient and extendable diagnostic models. 
To address these issues, we propose UniCAD, a unified architecture that leverages the robust capabilities of pre-trained vision foundation models to seamlessly handle both 2D and 3D medical images while requiring only minimal task-specific parameters. 
UniCAD introduces two key innovations:
(1) Efficiency: A low-rank adaptation strategy is employed to adapt a pre-trained visual model to the medical image domain, achieving performance on par with fully fine-tuned counterparts while introducing only 0.17\% trainable parameters.
(2) Plug-and-Play: A modular architecture that combines a frozen foundation model with multiple plug-and-play experts, enabling diverse tasks and seamless functionality expansion.
Building on this unified CAD architecture, we establish an open-source platform where researchers can share and access lightweight CAD experts, fostering a more equitable and efficient research ecosystem. 
Comprehensive experiments across 12 diverse medical datasets demonstrate that UniCAD consistently outperforms existing methods in both accuracy and deployment efficiency. 
The source code and project page are available at \url{https://mii-laboratory.github.io/UniCAD/}.
\end{abstract}


\begin{highlights}
\item \textbf{Unified Processing Across Dimensions}: The Unified Embedding Layer (UEL) in UniCAD enables seamless 2D/3D image processing within a shared Vision Foundation Model, enhancing 3D capabilities.
\item \textbf{Efficient Multi-Task Adaptation}:  UniCAD leverages 0.17\% low-rank adaptation (LoRA) for multi-task processing, maintaining single-task efficiency and full fine-tuning performance.
\item \textbf{Open Ecosystem for Collaborative CAD Development}: UniCAD’s open ecosystem allows minimal parameter updates and weight sharing, driving collaborative, scalable CAD development.
\end{highlights}



\maketitle


\section{Introduction}
\label{sec1}
\begin{figure}[t]
    \centering
    \includegraphics[width=0.48\textwidth]{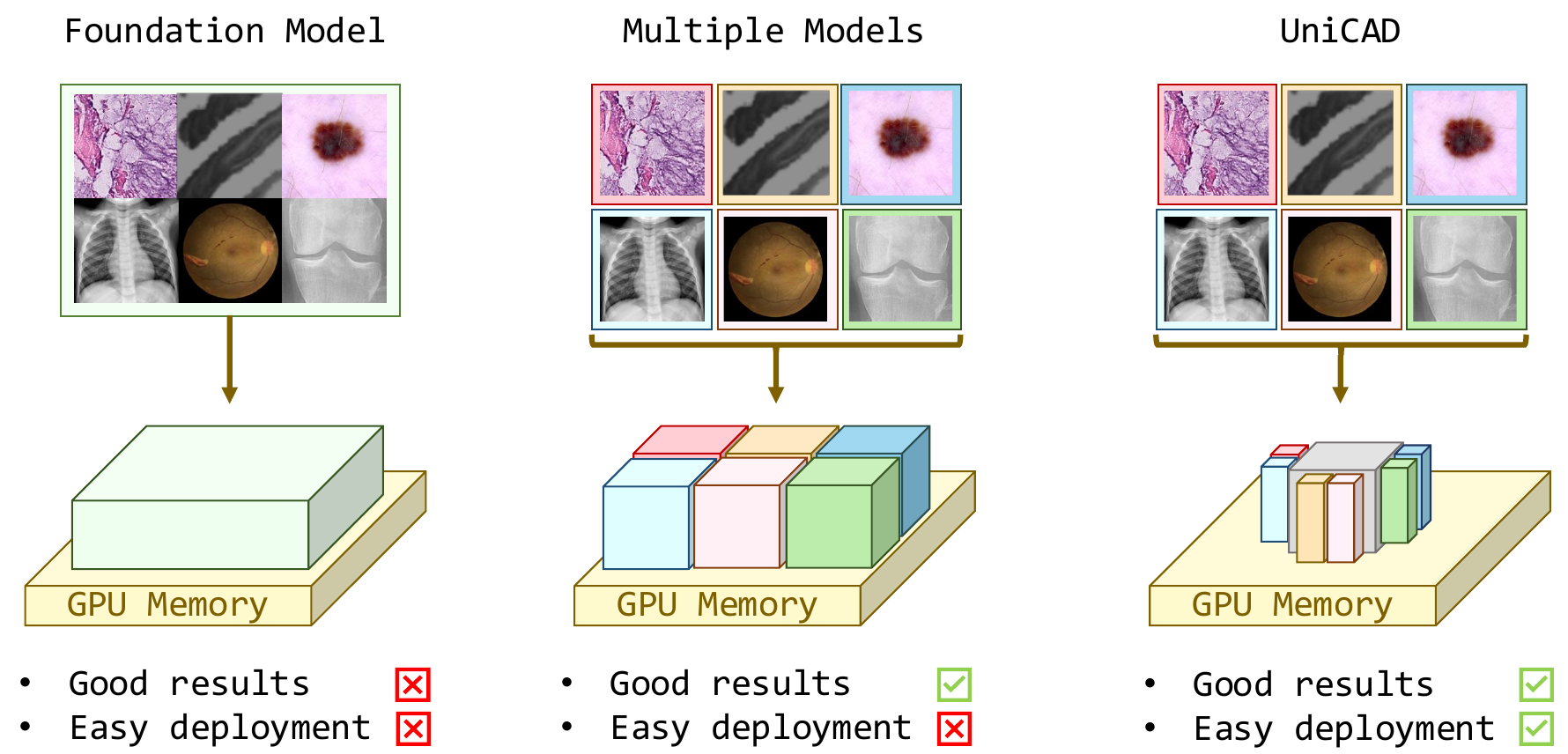}
    \caption{Comparison of different multi-task CAD systems. From left to right: (1) a single model handling multiple tasks, which consumes significant GPU memory and has limited deployment flexibility; (2) multiple independent models for distinct tasks, resulting in high memory consumption and redundancy; and (3) our proposed UniCAD, a unified architecture that combines a single general model with multiple small, task-specific experts, optimizing memory efficiency while maintaining flexibility for diverse tasks.
    }
    \label{teaser}
\end{figure}

Recent advancements in computer-aided diagnosis (CAD) systems for diverse medical images and clinical applications have been profoundly shaped by the evolution of deep learning models. 
As depicted in Figure \ref{teaser}, these modern multi-task CAD systems can be broadly classified into three main categories. 
The first category involves the development of a single model designed to handle all CAD tasks concurrently. 
The second category emphasizes constructing and deploying independent models tailored to each specific task. 
The third category takes advantage of a pre-trained vision foundation model that serves as a common image encoder, while only small trainable modules are added to address different CAD tasks. 

\begin{figure}[t]
    \centering
    \includegraphics[width=0.48\textwidth]{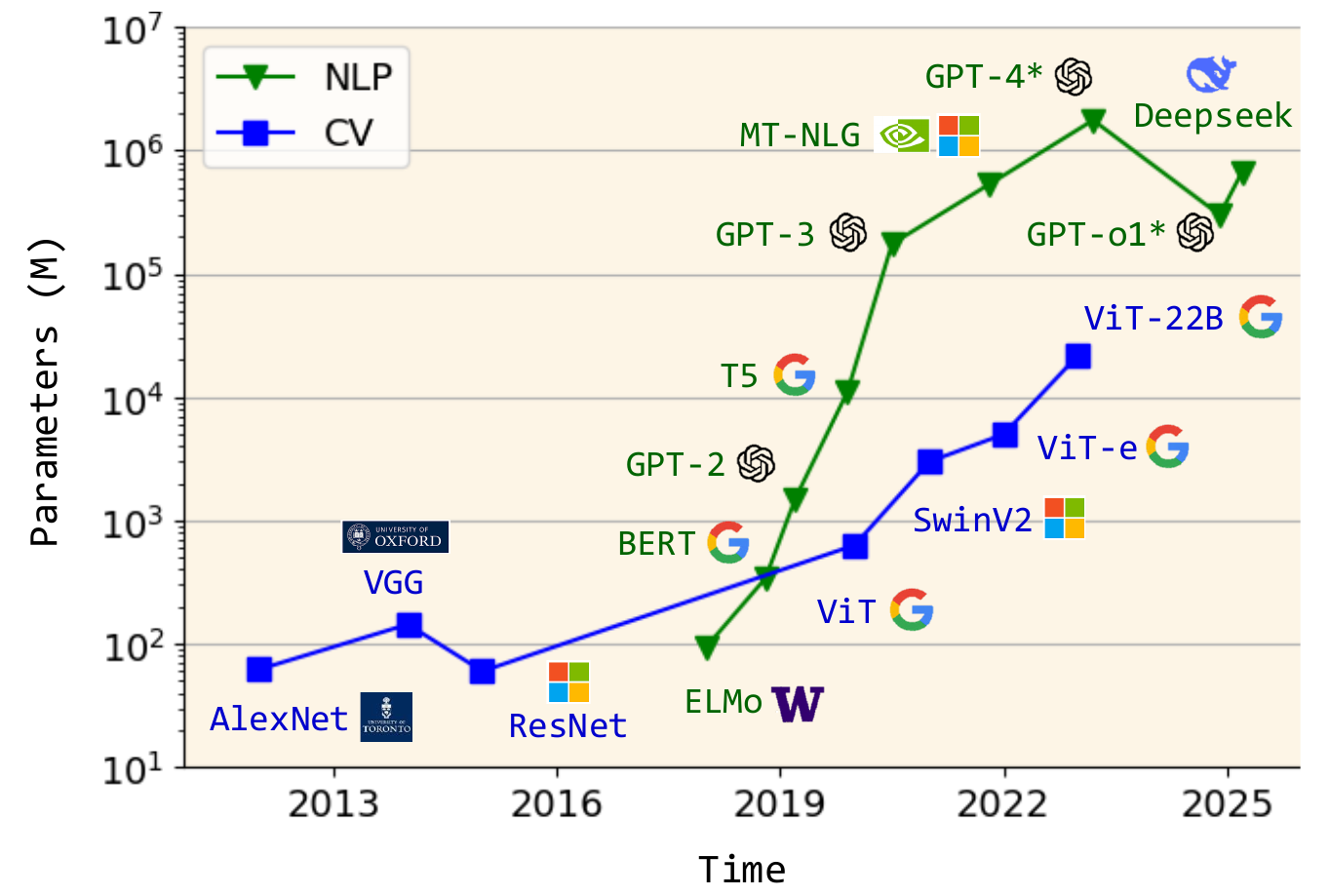}
    \caption{Recent trends have shown that the parameter sizes of models in both CV and NLP are rapidly increasing, achieving outstanding performance. This underscores the importance of leveraging these vision foundation models to enhance the capabilities of multi-task CAD systems. 
    ``*'' means the parameter size is an estimate, rather than the official release.}
    
    \label{nlpcv}
\end{figure}

The first category of multi-task CAD systems focuses on utilizing a single foundation model to handle all CAD tasks concurrently. 
This approach is appealing, as foundation models have profoundly impacted natural language processing (NLP) and computer vision (CV) over the last decade, with techniques from these fields directly informing advances in medical image analysis and diagnostics~\citep{dosovitskiy2020image,dehghani2023scaling}.
As illustrated in Figure \ref{nlpcv}, one key feature of vision foundation models, such as the Vision Transformer (ViT)~\citep{dosovitskiy2020image}, is their adherence to the scaling law observed in NLP models, indicating that model performance markedly improves with larger model sizes and data scales~\citep{dehghani2023scaling}.
However, training these models from scratch demands extensive and well-annotated datasets, which present significant challenges in the healthcare sector due to privacy concerns in data collection and the substantial costs associated with data annotation~\citep{razzak2018deep,shen2023image}. 
Additionally, a single model for multi-task CAD prioritizes generalization over task specialization~\citep{razzak2018deep}. This approach is problematic because medical imaging data is extremely heterogeneous. It spans various modalities (X-ray, CT, MRI, ultrasound), anatomical regions, pathologies, and institutional protocols. Each of these requires distinct diagnostic expertise and feature attention.
Moreover, the constantly evolving clinical demands require models to be frequently fine-tuned, risking catastrophic forgetting and demanding considerable GPU resources, which further complicates deployment and reduces system flexibility.

The second type of multi-task CAD system has evolved to address the limitations inherent in using a single model for all tasks. This approach focuses on constructing independent models specifically tailored for each clinical task. 
One common implementation approach involves fine-tuning pre-trained models for specific clinical tasks.
These pre-trained models originate from either general-purpose vision models trained on natural images or specialized models pre-trained on medical tasks.
These models can then be distilled into smaller, more efficient versions that maintain diagnostic accuracy while requiring less computing power.
Although this approach theoretically enables efficient memory management by loading only task-specific models when needed, it presents practical deployment challenges. 
The significant loading time when switching between models introduces operational latency that impacts clinical workflow, creating a difficult trade-off between memory efficiency and system responsiveness. 
Furthermore, when multiple models must remain loaded simultaneously to maintain responsiveness, significant storage redundancy occurs as each model contains largely overlapping architectural components, increasing infrastructure costs and limiting scalability in resource-constrained environments.

The third category of multi-task CAD systems, exemplified by frameworks like MeLo~\citep{zhu2024melo} and Reprogramming Distillation~\citep{zhou2024reprogramming}, has been proposed to address existing challenges. 
These approaches typically utilize a pre-trained vision foundation model as a shared image encoder. 
This encoder remains frozen, while only small, trainable modules are introduced for task-specific adaptations without requiring full parameter fine-tuning.
Such methods demonstrate excellent task extensibility by significantly reducing computational costs and preventing catastrophic forgetting when adapting to new CAD tasks during training.
From a deployment perspective, this architecture is particularly advantageous as it maintains a single shared backbone across all tasks, eliminating redundant model loading and enabling efficient multi-task switching with minimal memory overhead.
However, these methods are currently constrained to 2D medical images and have been validated on a limited number of CAD tasks, leaving their effectiveness across a broader spectrum of medical imaging modalities unverified.

\begin{figure*}[t]
    \centering
    \includegraphics[width=1\textwidth]{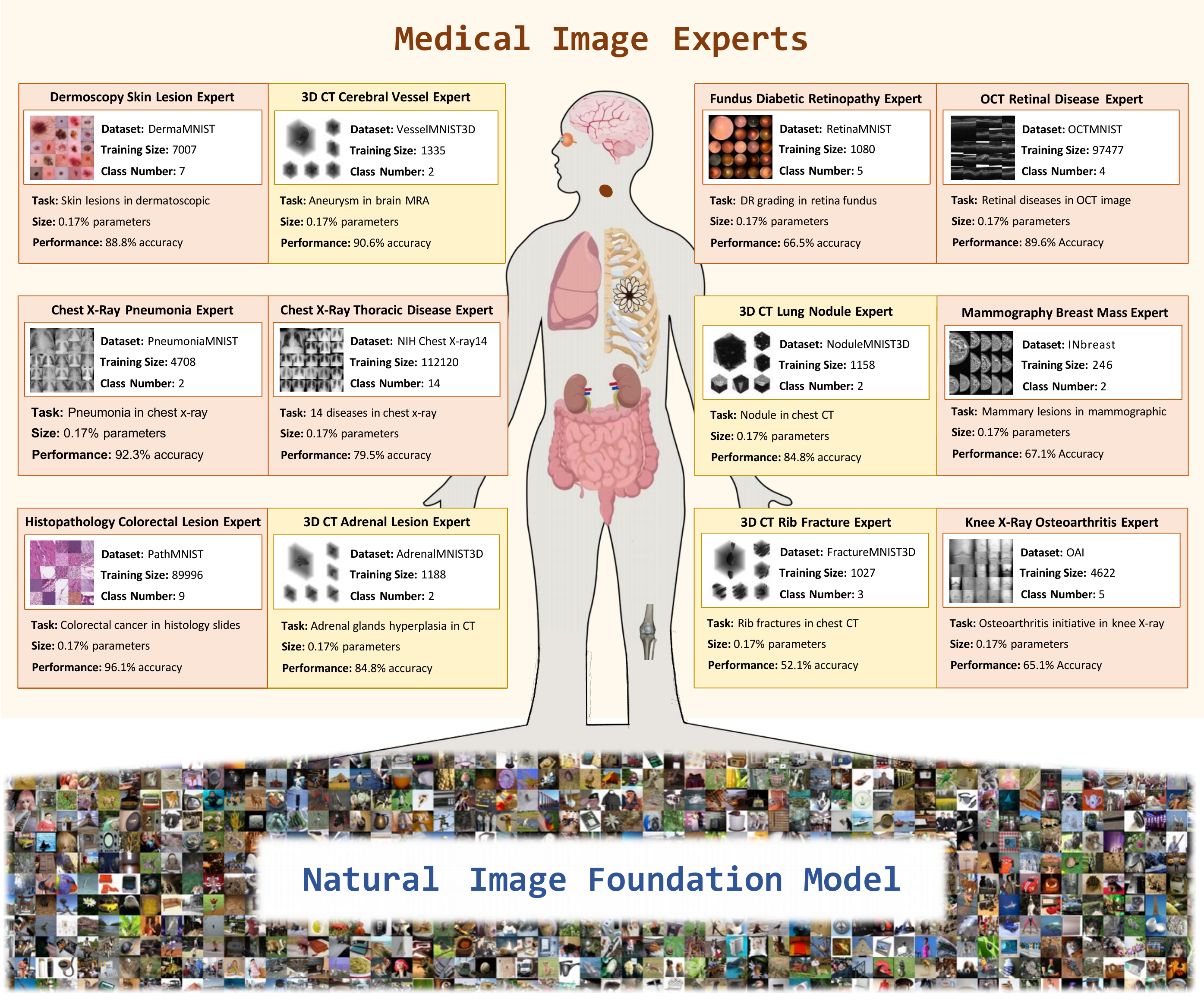}
    \caption{The visualization showcases the datasets and task-specific experts used in our experiments. Red boxes represent 2D datasets, while yellow boxes denote 3D datasets. Building on foundation models derived from natural images, we achieve a wide range of diagnostic tasks for the human body at a relatively low cost, while maintaining strong performance across tasks.
    }
    \label{dataset}
\end{figure*}

To overcome the challenges faced by existing multi-task CAD systems, we propose UniCAD, a unified solution that efficiently solves multiple CAD tasks through a single foundation model architecture. 
This work builds upon our previous MeLo approach~\citep{zhu2024melo}, which employs low-rank adaptation (LoRA)~\citep{hulora} to adapt a pre-trained vision foundation model to multiple CAD tasks with minimal parameter overhead. 
UniCAD creates a unified framework that seamlessly handles both 2D and 3D medical images, allowing users to submit diverse diagnostic tasks to a single system that intelligently manages all processing without requiring explicit model selection from users.
As illustrated in Figure~\ref{dataset}, UniCAD achieves state-of-the-art or competitive diagnostic performance across a diverse range of medical image datasets while maintaining exceptional task extensibility.
This positions UniCAD as a comprehensive solution for expanding the clinical applicability of multi-task CAD systems across varied medical imaging contexts.

Beyond technical performance, UniCAD enables a collaborative ecosystem for CAD development and deployment in future.
The architecture's design allows medical institutions to update existing diagnostic capabilities or deploy new tasks with only minimal parameter exchanges (0.17\% of total weights), based on a selected general-purpose vision foundation model.
Since the foundation model remains frozen across all deployments, institutions only need to transfer lightweight task-specific parameters, enabling efficient knowledge sharing while preserving patient data privacy.
This parameter-sharing approach allows diagnostic tasks to run flexibly across multiple datasets and clinical environments, fostering an open ecosystem where researchers can collaboratively advance CAD capabilities globally while maintaining stringent data protection standards.

In summary, the key contributions of UniCAD are as follows:
\begin{itemize}    
    \item \textbf{Unified 2D/3D Processing with a Vision Foundation Model.} Our architecture seamlessly integrates both 2D and 3D medical images through a unified embedding layer and shared backbone, enabling consistent diagnostic capabilities across different image dimensions.
    \item \textbf{Efficient Multi-Task Processing with Compact Experts.} By incorporating just 0.17\% additional low-rank weights to the vision foundation model, we achieve flexible task adaptation while maintaining performance and speed comparable to specialized single-task models.
    \item \textbf{Open Ecosystem for Enhanced CAD Collaboration.} We initiate a collaborative platform where institutions can share lightweight task-specific parameters without exchanging sensitive data, fostering global advancement of CAD capabilities while maintaining stringent privacy standards.
\end{itemize}

\section{Related Works}

\subsection{Medical Image Foundation Models}

In the field of medical image analysis, early approaches focused on task-specific models designed for particular applications~\citep{liu2020deep,mckinney2020international,heker2020joint}. 
Later, researchers discovered the potential of large models, leading to the development of a wide variety of medical image foundation models. CLIP-based models~\citep{radford2021learning,wang2022medclip,koleilat2024medclip} combine visual and language capabilities to enable powerful cross-modal understanding. 
One notable example is BiomedCLIP~\citep{zhang2024biomedclip}, which leverages a hybrid architecture of ViT-L/16 and PubMedBERT pretrained on 15 million biomedical image-text pairs. 
This model achieves exceptional performance in cross-modal retrieval and biomedical image classification tasks. 
Another significant contribution is PMC-CLIP~\citep{lin2023pmc}, which is trained on 1.6 million biomedical image-text pairs from PubMed Central. Using a CLIP-style architecture, PMC-CLIP demonstrates strong performance across various downstream tasks, including image-text retrieval, image classification, and medical visual question answering (VQA).

Building upon these vision-language approaches, another category of foundation models integrates large language models to enhance medical image analysis capabilities ~\citep{lai2024residual,li2023lvit,wu2023medklip}. 
For example, Med-Flamingo~\citep{moor2023med} leverages large language models by incorporating multimodal image-text data, enabling few-shot learning for complex medical VQA tasks. This approach allows the model to not only understand medical images but also generate clinically relevant responses, significantly enhancing its performance in tasks like diagnosis and rationale generation. Similarly, LLaVA-Med~\citep{li2023llava} adapts a general-domain vision-language model for the biomedical domain through curriculum learning, fine-tuning it with biomedical image-text pairs and GPT-4-generated instruction-following data to achieve state-of-the-art performance on biomedical VQA tasks. M3D-LaMed~\citep{bai2024m3d} further integrates large language models with 3D medical image encoders, achieving state-of-the-art performance in tasks like image-text retrieval, report generation, and 3D segmentation.
Additionally, several SAM~\citep{kirillov2023segment}-based foundation models have been developed for general medical image segmentation tasks~\citep{ma2024segment,wang2023sam,shaharabany2023autosam}. 
This rich variety of medical image foundation models has greatly facilitated advances in medical image processing and analysis.

Despite these impressive advances in medical foundation models, several practical challenges remain for clinical deployment. 
While these general-purpose foundation models demonstrate strong generalization capabilities across various tasks, they require immense computational resources and large-scale datasets for training—resources that are often limited in healthcare settings. 
Additionally, their performance on specialized clinical tasks frequently falls short of task-specific models, as the emphasis on broad generalization compromises precision in narrow diagnostic contexts. 
For instance, while Med-Flamingo~\citep{moor2023med} excels at few-shot learning and general visual reasoning, it struggles with fine-grained classification tasks critical for accurate diagnoses. 
This performance gap highlights the ongoing tension between general-purpose foundation models and the specialized, task-specific models needed for precise clinical applications.

\subsection{Task Adaptation of Pre-trained Models}

To circumvent the data scarcity issue, many works adapt models pre-trained on natural images (e.g., ResNet50~\citep{he2016deep} and ViT~\citep{dosovitskiy2020image}) to medical diagnosis tasks via fine-tuning. 
While full fine-tuning achieves strong performance, it risks catastrophic forgetting when adapting to new tasks and incurs prohibitive GPU memory costs during multi-task deployment.
Alternatively, linear probing offers a more parameter-efficient approach. In this method, the pre-trained backbone network is completely frozen, and only a new task-specific linear classification layer is trained on top of the extracted features. While this drastically reduces the number of trainable parameters and memory requirements, it often sacrifices accuracy compared to full fine-tuning, as evident in Table \ref{integration}.
Another common strategy involves training multiple independent models for different tasks, but it introduces redundancy in both storage and computation. 
For example, switching between task-specific models during inference exacerbates latency and hinders scalability. 
These approaches highlight a critical trade-off: while adapting pre-trained models can leverage prior knowledge, it struggles to achieve a balance between efficiency, flexibility, and performance in multi-task settings.

To address these challenges, recent studies have focused on parameter-efficient fine-tuning methods. 
Among these approaches, LoRA has emerged as a promising technique due to its ability to significantly reduce the number of trainable parameters while maintaining competitive performance.
Several methods have successfully applied LoRA to medical imaging tasks. 
In the segmentation domain, SAMed~\citep{zhang2023customized} fine-tunes the Segment Anything Model (SAM) for medical image segmentation. 
With only 5.25\% of parameters updated, it achieves 81.88 DSC on the Synapse dataset. 
Similarly, CPC-SAM~\citep{miao2024cross} enhances SAM for semi-supervised segmentation using cross prompting and prompt consistency. 
It reaches 85.56 DSC on the ACDC dataset with just one labeled sample.
For diagnosis tasks, MoRA~\citep{shi2024mora} introduces modality-aware LoRA to improve multi-modal disease diagnosis. 
Using only 1.6\% trainable parameters, it outperforms previous methods across various diagnostic benchmarks. 
These examples demonstrate that LoRA can effectively adapt foundation models to specialized medical tasks with minimal parameter updates.
Moreover, recent research has extended this concept to multi-task learning scenarios with MOELoRA~\citep{liu2023moelora}, which combines Mixture-of-Experts (MOE) with LoRA for efficient multi-task learning.

Building on these parameter-efficient approaches, recent advances have extended LoRA's application to complex multi-task scenarios in medical imaging.
MeLo~\citep{zhu2024melo} introduces a LoRA-based approach to efficiently fine-tune pre-trained vision models across diverse diagnostic tasks while maintaining high performance.
By freezing the backbone and injecting minimal trainable parameters for each task, MeLo achieves impressive efficiency, avoids catastrophic forgetting, and reduces computational overhead.
In this work, UniCAD will further develop a novel architecture that supports both 2D and 3D modalities through a unified ViT backbone with innovative dimensionality-adaptive embeddings.
Unlike methods such as M3D-LaMed~\citep{bai2024m3d}, which require specialized 3D pre-training, UniCAD leverages the rich visual representations from natural image pre-training to achieve state-of-the-art efficiency across diverse medical tasks while significantly reducing computational costs and enabling more accessible deployment in resource-constrained environments.
Additionally, the 2D natural image pre-training used by UniCAD is much more widely available and easier to access than specialized 3D pre-training datasets, further enhancing its practical applicability in diverse clinical settings.

\section{Method}
\label{sec4}
In this section, we present our methodology, which leverages the strengths of pre-trained models while enabling efficient fine-tuning for specialized tasks. 
Specifically, our framework is capable of processing both 2D and 3D images, along with the deployment paradigm designed to handle flows of diverse tasks requesting service in random orders, enhancing its adaptability to real-world clinical deployment challenges. The following subsections provide a detailed exposition of our approach.



\subsection{Overall Architecture}

\begin{figure*}[t]
    \centering
    \includegraphics[width=1\textwidth]{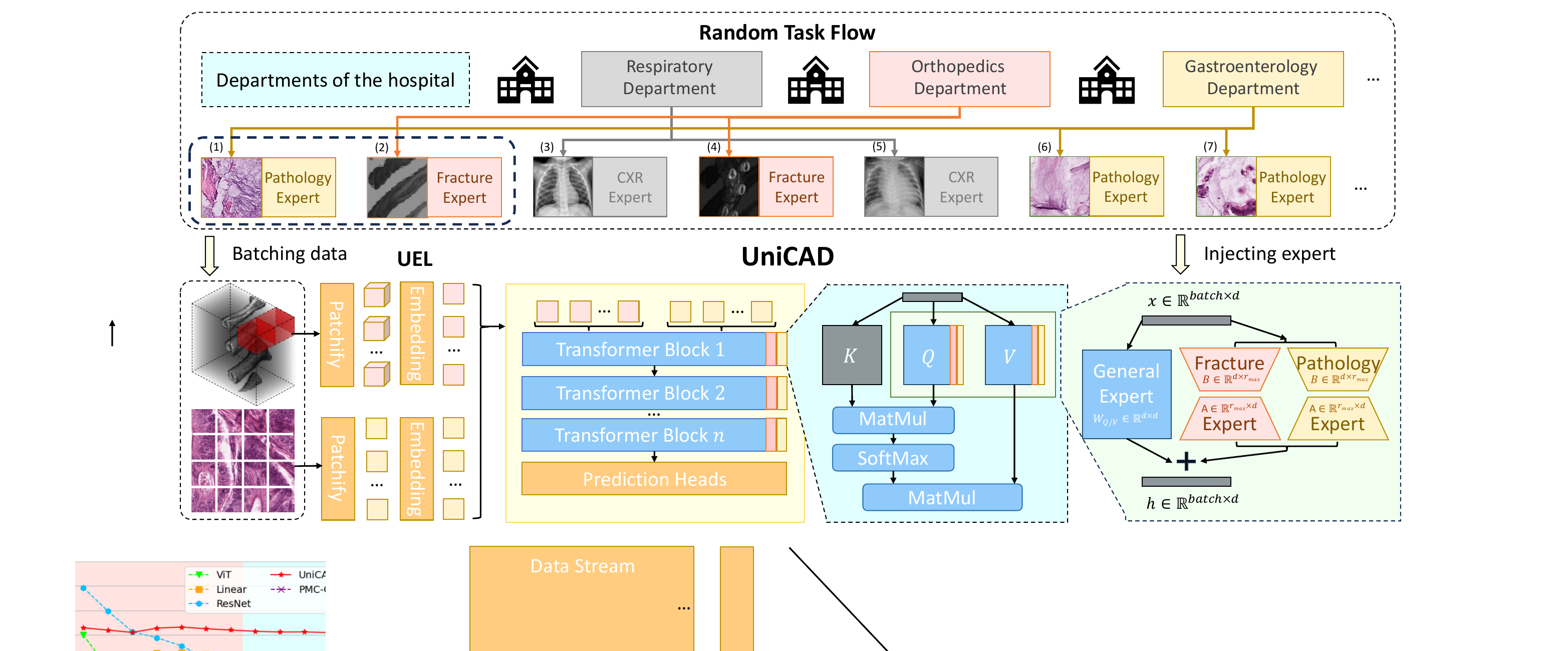}
    \caption{
    The illustration of overall architecture depicts how images are handled through UniCAD, using 3D fracture, 2D pathology, and chest X-ray as examples in the random task flow. 
    At its core, UniCAD employs a Unified Embedding Layer (UEL) that seamlessly integrates both 2D and 3D medical images within a single framework. 
    This architecture enables adaptable and versatile diagnostic capabilities for diverse clinical applications.
}
    \label{method}
\end{figure*}

Our UniCAD system is designed to deliver superior performance in multi-task diagnostic deployments, providing high flexibility and efficient use of hardware resources.
As shown in Figure \ref{method}, we simplify the diverse and unordered diagnostic requests in medical centers as a \textit{random task flow}.
This flow is flexible and allows dynamic adaptation to various tasks, as UniCAD can process both 2D and 3D medical images through specialized embedding layers.
These embedding layers standardize images into a unified sequence of feature tokens, ensuring compatibility across different image types.
Once standardized, the system efficiently batches the data and tasks for parallel processing, enabling low-latency execution across multiple diagnostic requests.
In the Transformer blocks for diagnosis, expert weights are dynamically injected based on the specific task, ensuring accurate and efficient processing.
In general, our architecture enables UniCAD to deliver precise diagnostic results across multiple tasks and imaging modalities while optimizing resource utilization, eliminating the need for specialized hardware and making it an ideal solution for real-world clinical environments.

\subsection{Unified Embedding Layer}


To handle both 2D and 3D medical images within a single framework, we introduce the Unified Embedding Layer (UEL). This layer converts inputs of varying spatial dimensions into a standardized token sequence compatible with the ViT backbone. By using this unified embedding strategy, we enable the seamless integration of mixed 2D and 3D inputs, allowing the framework to support a wide range of medical imaging tasks without requiring modality-specific adjustments. 

\textbf{Image Embedding}: The embedding strategy for both 2D and 3D images follows a similar approach, with necessary modifications to accommodate their dimensional differences. For a 2D image \( x \in \mathbb{R}^{H \times W \times C} \), the image is divided into patches of size \( P_H \times P_W \), and each patch is flattened and linearly projected into a \( d \)-dimensional token using a trainable projection matrix \( E \in \mathbb{R}^{(P_H \cdot P_W \cdot C) \times d} \). For a 3D volumetric image \( I \in \mathbb{R}^{D \times H \times W \times C} \), the image is partitioned into volumetric patches of size \( P_D \times P_H \times P_W \), and then patchified into \( d \)-dimensional tokens using a projection matrix \( E_v \in \mathbb{R}^{(P_D \cdot P_H \cdot P_W \cdot C) \times d} \). 

For spatial awareness, we incorporate positional embeddings. For 2D images, the positional embeddings $E_{\text{pos}} \in \mathbb{R}^{(N + 1) \times d}$ are based on the 2D patch count $N = \frac{H \cdot W}{P_{H}\times P_{W}}$, while for 3D images, the embeddings $E_{\text{pos}} \in \mathbb{R}^{(N + 1) \times d}$ are derived from the volumetric patch count $N = \frac{D \times H \times W}{P_D \times P_H \times P_W}$.
In both cases, a learnable classification token (CLS) $t_{\text{cls}} \in \mathbb{R}^d$ is prepended to the sequence of patch tokens to capture global semantic features through self-attention mechanisms. 
The above unified embedding strategy ensures that both 2D and 3D images are processed in a consistent format, enabling the model to handle diverse medical imaging tasks efficiently.



\textbf{Batch Standardization}: 
For task requests that arrive within a short timeframe, we can batch them together rather than processing them sequentially, significantly improving throughput and resource utilization.
To support efficient processing across a batch of heterogeneous inputs, we perform batch standardization to align both 2D and 3D images into the same Transformer-compatible format.
After processing through the UEL, both 2D and 3D images are converted into token sequences with potentially varying lengths due to differences in original image dimensions, making them dimensionally compatible regardless of their original format.
To standardize these sequences, we apply padding to match a pre-defined maximum sequence length.
A binary masking mechanism ensures that only valid tokens contribute to the subsequent computation in Transformer blocks, with zeros assigned to padded positions and ones to actual image content.
This standardization approach enables UniCAD to seamlessly process diverse imaging data without requiring modality-specific adaptations, maintaining both computational efficiency and architectural elegance.

\subsection{Medical Image Low-rank Adaptation}
To efficiently adapt a ViT model to specific CAD applications, we utilize the parameter-efficient fine-tuning technique known as LoRA~\citep{hulora}. Originally proposed for fine-tuning large language models, LoRA is based on the observation that weight changes during fine-tuning of pre-trained large models are sparse and exhibit a low intrinsic rank. The core idea of LoRA is to freeze the original weights of the pre-trained model, and introduce trainable low-rank decomposition matrices into each layer of the Transformer architecture. This approach significantly reduces the number of trainable parameters required when fine-tuning large models.

In UniCAD, we leverage LoRA to efficiently adapt ViT-based models for various diagnostic tasks. By freezing the pre-trained Transformer weights, which serve as the \textit{general expert}, and incorporating LoRA weights as \textit{domain-specific experts}, we tailor the model’s response to the specific nuances of different medical imaging tasks. 
To achieve this, we integrate LoRA weights into the self-attention layers of a pre-trained ViT model, as illustrated in Figure~\ref{method}.
These LoRA weights specifically modify the query and value projection matrices ($W_Q$ and $W_V$) within the self-attention layers in all Transformer blocks.
The modification is accomplished by constraining the updates of these matrices through a low-rank decomposition during the fine-tuning process.
The mathematical representation of this modification is as follows:
\begin{equation}
h_{j} = W_{0}x_{j} + \Delta W x_{j} = W_{0}x_{j} + B A x_{j},
\label{eq-1}
\end{equation}
where \( x_{j} \in \mathbb{R}^{d} \) represents the \( j \)-th token of the input feature sequence, and \( h_{j} \in \mathbb{R}^{d} \) denotes the \( j \)-th token of the layer's output.
The weight adjustment \( \Delta W \) is expressed as the product of two low-rank matrices, \( B \in \mathbb{R}^{d \times r} \) and \( A \in \mathbb{R}^{r \times d} \), which are applied to the pre-trained weight \( W_{0} \).
For each task, these expert matrices ($A$ and $B$) have rank \( r \), which is substantially smaller than the token dimension \( d \), allowing for efficient updates while maintaining the model's ability to robustly handle multiple tasks.

\subsection{Multi-Task Batch Processing}

To support clinical deployments that handle a flow of diverse tasks, we process multiple tasks concurrently within a batch of size \( n \).
For the $i$-th task, the input to a Transformer block is a sequence of tokens $\{x_{ij}\}$, and the output is $\{h_{ij}\}$, where $i$ indexes the task in the batch and $j$ indexes the tokens.
The index $j$ ranges from 1 to $N+1$, with $N$ being the number of patch tokens and the additional token being the CLS token.
While each Transformer block processes this CLS token along with patch tokens, only the CLS token from the final Transformer block in the network architecture is used for the final image classification.

Following equation \eqref{eq-1}, we first compute $Ax$ and then multiply the result with $B$.
Initially, each token $x_{ij}$ should maintain consistent dimensionality $d$ across all tasks and patches in the batch.
However, when applying the task-specific $A$ matrices, the resulting intermediate outputs have dimensions that vary across tasks, since the low-rank value \( r \) can differ between domain-specific experts.
This dimensional variance occurs because the expert matrices $A \in \mathbb{R}^{r \times d}$ and $B \in \mathbb{R}^{d \times r}$ are independently learned for each task and can even be developed by different researchers.
These dimensional inconsistencies prevent efficient parallel computation within the batch, as modern GPU accelerators require uniform tensor shapes for optimal batch processing.

To address this dimensional inconsistency challenge, we implement a rank standardization approach that enables efficient parallel processing across diverse tasks.
Our solution identifies the maximum rank value $r_{\text{max}}$ across all tasks in the current batch and standardizes matrix dimensions through zero-padding.
Specifically, for any task where the native rank $r$ is smaller than $r_{\text{max}}$, we pad the matrices $B$ and $A$ with zeros along their low-rank dimension.
This zero-padding technique is conceptually similar to our earlier token sequence padding strategy, as both methods standardize tensor shapes to enable efficient batch processing.
The padding ensures that all domain-specific experts have uniform dimensions ($B^{\prime} \in \mathbb{R}^{d \times r_{\text{max}}}$ and $A^{\prime} \in \mathbb{R}^{r_{\text{max}} \times d}$) throughout the batch.
With these standardized matrices, we can express the forward computation for each token $x_{ij}$ as:
\begin{equation}
h_{ij} = W_{0} x_{ij} + B_{i}^{\prime} A_{i}^{\prime} x_{ij}.
\end{equation}
This formulation enables each task-specific adaptation, represented by the respective pairs of matrices $B_i^{\prime}A_i^{\prime}$, to be applied efficiently to the corresponding inputs while standardizing tensor dimensions for optimal hardware acceleration during parallel task processing.

At the final stage of our diagnostic pipeline, we leverage the CLS token produced by the last Transformer block to generate task-specific predictions.
This token is processed through a task-specific linear classification head to generate class predictions for the corresponding diagnostic task.
During training, standard cross-entropy loss computed between predicted probabilities and ground truth labels is backpropagated to update both the task-specific classification head and corresponding LoRA weights, while keeping the foundation model parameters frozen.
This linear probing approach maintains parameter efficiency while allowing each task to develop specialized diagnostic capabilities by isolating trainable parameters to the classification heads and LoRA weights, preventing interference between tasks and ensuring model updates remain focused on task-relevant features.
\subsection{Implementation Details}
In our research, all models are independently trained using an initial learning rate of 3e-4 with the Adam optimizer. These models are based on Vision Transformers (ViTs) and are trained for 100 epochs until convergence. We select the weights that demonstrate the best performance on validation data as our final model for testing. For Experiment \ref{exp1}, we utilize a ViT-based model pre-trained on the ImageNet dataset. For the 3D dataset, we utilize the 3D image embedding layer from UEL along with ViT for processing. In another experiment, we test ViTs of various sizes (\textit{base}, \textit{huge}, \textit{giant}, and \textit{giga}) pre-trained using CLIP, as referenced in ~\citep{radford2021learning}. Specifically, in Experiment \ref{exp2}, we evaluate a ViT-giga model pre-trained via CLIP. The pre-trained weights for all ViT models are obtained from the source ~\citep{rw2019timm}. All tests are conducted on a single Nvidia A100 GPU with 80GB of memory.

\section{Experiments}
To rigorously evaluate UniCAD, we conduct systematic experiments across three critical dimensions: (1) diagnostic accuracy on diverse 2D/3D medical imaging tasks, (2) computational efficiency under multi-task deployment scenarios, and (3) scalable performance with progressively larger foundation models while maintaining parameter efficiency. We first detail our experimental datasets and implementation protocols. Subsequent analyses validate UniCAD's diagnostic precision against state-of-the-art baselines, quantify its resource efficiency through our proposed DEAR metric, and demonstrate how it effectively leverages scaled foundation models without proportional parameter growth.

\subsection{Datasets}

To comprehensively evaluate the effectiveness of UniCAD, we conduct experiments on a diverse collection of medical imaging datasets, comprising eight 2D datasets and four 3D datasets.
These datasets represent 12 distinct diagnostic tasks across multiple anatomical regions, imaging modalities, and pathologies, which are illustrated in Figure \ref{dataset}.
The 2D datasets include thoracic disease classification from chest X-rays (NIH Chest X-ray14), tissue classification from histology images (PathMNIST), skin lesion identification (DermaMNIST), retinal disease diagnosis from OCT images (OCTMNIST), pneumonia detection (PneumoniaMNIST), diabetic retinopathy grading (RetinaMNIST), breast malignancy assessment (INbreast), and knee osteoarthritis staging (OAI).
The 3D datasets consist of lung nodule classification (NoduleMNIST3D), adrenal gland abnormality detection (AdrenalMNIST3D), rib fracture identification (FractureMNIST3D), and brain vessel aneurysm diagnosis (VesselMNIST3D).
This comprehensive selection ensures rigorous evaluation of UniCAD's unified architecture across varied dimensionality, resolution, and clinical contexts, while also demonstrating the potential for our approach to extend beyond these experimental settings to a wider range of real-world clinical applications.
Below, we provide detailed descriptions of each dataset and its corresponding diagnostic task.

\subsubsection*{2D Datasets}


\textbf{NIH Chest X-ray14}~\citep{wang2017chestx} is a comprehensive dataset containing 112,120 frontal-view chest radiographs.
Each image is annotated with one or more of 14 common thoracic diseases, forming a multi-label classification task where chest radiographs may contain multiple disease labels simultaneously.
Following the official guidelines, we divided the dataset into 75,708 samples for training, 10,816 samples for validation, and 25,596 samples for testing.
The dataset serves as a benchmark for automated diagnosis of thoracic pathologies from radiographic images.

\textbf{MedMNIST2D}~\citep{medmnistv2} consolidates a variety of medical imaging datasets for specific classification tasks.
The subsets used in the experiments are as follows:
\begin{itemize}
    \item 
    \textbf{PathMNIST} contains colorectal cancer histology images for tissue classification, comprising 9 distinct tissue types.
    This dataset consists of 89,996 training samples, 10,004 validation samples, and 7,180 test samples.
    \item 
    \textbf{DermaMNIST} features dermatoscopic images from the HAM10000 dataset for identifying skin lesions, divided into 7 lesion types.
    There are 7,007 samples allocated for training, 1,003 for validation, and 2,005 samples in the test set.
    \item
    \textbf{OCTMNIST} includes OCT images for diagnosing retinal diseases, standardized across the dataset, with 4 categories of retinal conditions.
    The dataset consists of 97,477 samples for training, 10,832 for validation, and 1,000 samples for testing.
    \item  
    \textbf{PneumoniaMNIST} presents pediatric chest X-rays for pneumonia classification, uniformly scaled, with 2 categories: pneumonia and normal.
    This dataset includes 4,708 samples for training, 524 for validation, and 624 samples for testing.
    \item  
    \textbf{RetinaMNIST} comprises retinal images from the DeepDRiD challenge for diabetic retinopathy grading with 5 severity levels.
    The dataset is split into 1,080 training samples, 120 validation samples, and 400 test samples.    
\end{itemize}

\textbf{INbreast}~\citep{InsMoreira2012INbreastTA} includes 410 digital mammography images from 115 patients, comprising 339 non-malignant and 71 malignant cases.
The diagnosis task follows BI-RADS assessment of masses~\citep{LauraLiberman2002BreastIR} to classify these mammography images into 2 categories: non-malignant and malignant.
The dataset is split into 60\% (246 images) for training, 20\% (82 images) for validation, and 20\% (82 images) for testing.

\textbf{OAI}~\citep{kellgren1957radiological} contains X-ray images from a multi-center, longitudinal study focused on knee osteoarthritis.
The dataset includes 5,778 knee X-ray images illustrating 5 distinct stages of knee osteoarthritis severity.
These stages are classified according to the widely utilized Kellgren and Lawrence grading system, ranging from Grade 0 (normal) to Grade 4 (severe).
The dataset is split into 80\% (4,622 images) for training, 10\% (578 images) for validation, and 10\% (578 images) for testing.

\subsubsection*{3D Datasets}

\textbf{MedMNIST3D}~\citep{medmnistv2} consolidates diverse 3D medical imaging datasets tailored for specific diagnostic and segmentation tasks.
The subsets used in the experiments are as follows:
\begin{itemize}
    \item 
    \textbf{NoduleMNIST3D} is derived from the LIDC-IDRI database, focusing on classifying lung nodules into 2 categories: benign or malignant.
    The dataset includes 1,158 training samples, 165 validation samples, and 310 test samples.
    \item 
    \textbf{AdrenalMNIST3D} features 3D adrenal gland masks from CT scans for normal or mass classification, with 2 distinct categories.
    The dataset consists of 1,188 training samples, 98 validation samples, and 298 test samples.
    \item 
    \textbf{FractureMNIST3D} is derived from the RibFrac Dataset and includes uniformly resized images of rib fractures, classified into 3 types.
    The dataset is split into 1,027 training samples, 103 validation samples, and 240 test samples.
    \item 
    \textbf{VesselMNIST3D} is derived from the IntrA dataset and contains 3D brain vessel models from MRA images, segmented into healthy and aneurysm categories.
    The dataset consists of 1,335 training samples, 191 validation samples, and 382 test samples.
\end{itemize}


\subsection{Diagnostic Accuracy in Diverse Tasks}
\label{exp1}

To evaluate the effectiveness of UniCAD across diverse medical image diagnosis tasks, we conducted comprehensive experiments on 11 single-label classification tasks and one multi-label classification task (NIH Chest X-ray14).
Our evaluation framework systematically compares methods across the three architectural categories established in the introduction.
For the first category (single foundation models for all tasks), we compare with general visual-language models including PMC-CLIP~\citep{lin2023pmc}, BiomedClip~\citep{zhang2024biomedclip}, and Med-Flamingo ~\citep{moor2023med}, all pre-trained on large-scale medical datasets.
Since PMC-CLIP~\citep{lin2023pmc} and BiomedClip~\citep{zhang2024biomedclip} were not originally designed for 3D medical images, we adapted them by extracting features from individual 2D slices of each volume and averaging these features to represent the entire 3D image.
For 3D comparison in this category, we additionally include M3D-LaMed~\citep{bai2024m3d}, a specialized 3D foundation model pre-trained on large-scale medical image datasets, as Med-Flamingo~\citep{moor2023med} only supports 2D inputs.
For the second category (individual models for each task), we compare against independently fine-tuned ViT \textit{base} models and ResNet50 models(pre-trained on ImageNet with all parameters updated), representing the approach of deploying separate, task-specific models for each diagnostic task. For the 3D ViT, we use a combination of pre-trained 3D embedding layers and transformer block parameters pre-trained on ImageNet, while the 3D ResNet50 is fine-tuned using the pre-trained weights from PyTorch's official ResNet50 model.
For the third category (shared foundation model with task-specific adaptations), we present our UniCAD approach alongside a ViT with linear probing, which trains only the final classification head while keeping the backbone frozen.

\textbf{Single-Label Classification:}
The diagnostic accuracy for 11 single-label classification tasks is presented in Table \ref{integration}, demonstrating UniCAD's performance advantage in both ACC and F1 scores across 2D and 3D medical image diagnostic tasks.
In 2D comparison, fully fine-tuned ViT emerges as UniCAD's strongest competitor, followed by ResNet50 in overall performance.
Notably, general visual-language foundation models like BiomedCLIP and Med-Flamingo struggle to achieve strong performance in zero-shot inference.
To address the particularly poor zero-shot performance of PMC-CLIP, we fine-tuned the model on all 2D datasets in Table \ref{integration}.
This fine-tuning resulted in PMC-CLIP*, which shows improved but still suboptimal performance when handling diagnostic tasks across diverse modalities.
These results highlight the challenges of transferring general semantic understanding to specialized medical domains without significant fine-tuning on domain-specific data.
In 3D comparison, the results follow similar patterns as those observed in 2D tasks, with UniCAD achieving performance comparable to the fine-tuned 3D ViT while outperforming other methods.
In conclusion, UniCAD demonstrates robust and effective performance across both 2D and 3D diagnostic tasks, validating its versatility as a unified diagnostic framework.

\begin{table*}[h]
\caption{Comparative performance metrics of models across binary classification tasks in 2D and 3D. The \colorbox{mygray}{gray} section represents our proposed method. \textbf{Bold} indicates the best result, and the \underline{underlined} number denotes the second-best result.}


\renewcommand{\arraystretch}{1.3}
\centering

\resizebox{\textwidth}{!}{
\begin{tabular}{l|cccccc|cccc|cccc}
\hline
Model Category        & \multicolumn{6}{c|}{General Foundation Model}                                                                                                                             & \multicolumn{4}{c|}{Task-Specific Model}                                                  & \multicolumn{4}{c}{Unified Multi-Task Model}                                 \\ \hline
{\multirow{2}{*}{2D Tasks}} & \multicolumn{2}{c|}{\begin{tabular}[c]{@{}c@{}}PMC-CLIP* \\ \citep{lin2023pmc}\end{tabular}}      & \multicolumn{2}{c|}{\begin{tabular}[c]{@{}c@{}}BiomedCLIP \\ \citep{zhang2024biomedclip}\end{tabular}} & \multicolumn{2}{c|}{\begin{tabular}[c]{@{}c@{}}Med-Flamingo \\ \citep{moor2023med}\end{tabular}} & \multicolumn{2}{c|}{\begin{tabular}[c]{@{}c@{}}ViT (fine-tuned) \\ \citep{dosovitskiy2020image}\end{tabular}}                   & \multicolumn{2}{c|}{\begin{tabular}[c]{@{}c@{}}ResNet50 \\ \citep{he2016deep}\end{tabular}}    & \multicolumn{2}{c|}{\cellcolor[HTML]{EFEFEF}UniCAD}        & \multicolumn{2}{c}{ViT (linear probing)} \\
\multicolumn{1}{c|}{}             & ACC             & \multicolumn{1}{c|}{F1}             & ACC               & \multicolumn{1}{c|}{F1}                & ACC                       & F1                       & ACC             & \multicolumn{1}{c|}{F1}              & ACC            & F1              & \cellcolor[HTML]{EFEFEF}ACC             & \multicolumn{1}{c|}{\cellcolor[HTML]{EFEFEF}F1}              & ACC                 & F1                 \\ \hline
OCTMNIST              & 0.250           & \multicolumn{1}{c|}{0.100}          & 0.463             & \multicolumn{1}{c|}{0.354}             & 0.260                     & 0.235                    & \underline{0.893} & \multicolumn{1}{c|}{\textbf{0.896}}  & 0.884          & 0.885           & \cellcolor[HTML]{EFEFEF}\textbf{0.896}  & \multicolumn{1}{c|}{\cellcolor[HTML]{EFEFEF}\underline{0.895}} & 0.771               & 0.746              \\
PathMNIST             & 0.184           & \multicolumn{1}{c|}{0.078}          & 0.328             & \multicolumn{1}{c|}{0.213}             & 0.110                     & 0.102                    & \textbf{0.965}  & \multicolumn{1}{c|}{\textbf{0.954}}  & 0.930          & 0.900           & \cellcolor[HTML]{EFEFEF}\underline{0.961} & \multicolumn{1}{c|}{\cellcolor[HTML]{EFEFEF}\underline{0.943}} & 0.936               & 0.913              \\
DermaMNIST            & 0.316           & \multicolumn{1}{c|}{0.090}          & 0.533             & \multicolumn{1}{c|}{0.186}             & 0.148                     & 0.099                    & \underline{0.883} & \multicolumn{1}{c|}{\underline{0.816}} & 0.849          & 0.725           & \cellcolor[HTML]{EFEFEF}\textbf{0.888}  & \multicolumn{1}{c|}{\cellcolor[HTML]{EFEFEF}\textbf{0.824}}  & 0.821               & 0.633              \\
PneumoniaMNIST        & 0.625           & \multicolumn{1}{c|}{0.400}          & 0.587             & \multicolumn{1}{c|}{0.576}             & 0.380                     & 0.272                    & 0.877           & \multicolumn{1}{c|}{0.857}           & 0.907          & 0.895           & \cellcolor[HTML]{EFEFEF}\textbf{0.923}  & \multicolumn{1}{c|}{\cellcolor[HTML]{EFEFEF}\textbf{0.914}}  & \underline{0.915}     & \underline{0.905}    \\
RetinaMNIST           & 0.435           & \multicolumn{1}{c|}{0.121}          & 0.148             & \multicolumn{1}{c|}{0.085}             & 0.200                     & 0.170                    & \underline{0.652} & \multicolumn{1}{c|}{0.445}           & 0.647          & 0.456           & \cellcolor[HTML]{EFEFEF}\textbf{0.665}  & \multicolumn{1}{c|}{\cellcolor[HTML]{EFEFEF}\textbf{0.520}}  & 0.600               & \underline{0.468}    \\
INbreast              & \underline{0.671} & \multicolumn{1}{c|}{\textbf{0.595}} & 0.316             & \multicolumn{1}{c|}{0.240}             & 0.355                     & 0.329                    & \textbf{0.684}  & \multicolumn{1}{c|}{0.406}           & 0.658          & \underline{0.544} & \cellcolor[HTML]{EFEFEF}\underline{0.671} & \multicolumn{1}{c|}{\cellcolor[HTML]{EFEFEF}0.402}           & 0.632               & 0.446              \\
OAI                   & 0.420           & \multicolumn{1}{c|}{0.118}          & 0.239             & \multicolumn{1}{c|}{0.077}             & 0.176                     & 0.154                    & \underline{0.642} & \multicolumn{1}{c|}{0.553}           & 0.623          & \underline{0.570} & \cellcolor[HTML]{EFEFEF}\textbf{0.651}  & \multicolumn{1}{c|}{\cellcolor[HTML]{EFEFEF}\textbf{0.601}}  & 0.528               & 0.294              \\ \hline
                      \multirow{2}{*}{3D Tasks}& \multicolumn{2}{c|}{\begin{tabular}[c]{@{}c@{}}PMC-CLIP* \\ \citep{lin2023pmc}\end{tabular}}      & \multicolumn{2}{c|}{\begin{tabular}[c]{@{}c@{}}BiomedCLIP \\ \citep{zhang2024biomedclip}\end{tabular}} & \multicolumn{2}{c|}{\begin{tabular}[c]{@{}c@{}}M3D-LaMed \\ \citep{bai2024m3d}\end{tabular}}           & \multicolumn{2}{c|}{3D ViT (fine-tuned)}                            & \multicolumn{2}{c|}{3D ResNet50}    & \multicolumn{2}{c|}{\cellcolor[HTML]{EFEFEF}UniCAD}        & \multicolumn{2}{c}{3D ViT (linear probing)}            \\
\multicolumn{1}{c|}{}              & ACC             & \multicolumn{1}{c|}{F1}             & ACC               & \multicolumn{1}{c|}{F1}                & ACC                       & F1                       & ACC             & \multicolumn{1}{c|}{F1}              & ACC            & F1              & \cellcolor[HTML]{EFEFEF}ACC             & \multicolumn{1}{c|}{\cellcolor[HTML]{EFEFEF}F1}              & ACC                 & F1                 \\ \hline
VesselMNIST           & 0.887           & \multicolumn{1}{c|}{0.470}          & 0.126             & \multicolumn{1}{c|}{0.117}             & 0.778                   & 0.507                    & \underline{0.893} & \multicolumn{1}{c|}{0.623}           & \textbf{0.906} & \textbf{0.664}  & \textbf{\cellcolor[HTML]{EFEFEF}0.906}  & \multicolumn{1}{c|}{\cellcolor[HTML]{EFEFEF}\underline{0.653}} & 0.882               & 0.490              \\
NoduleMNIST           & 0.794           & \multicolumn{1}{c|}{0.442}          & 0.206             & \multicolumn{1}{c|}{0.171}             & 0.613           & 0.517          & \underline{0.842} & \multicolumn{1}{c|}{\textbf{0.763}}  & 0.839          & 0.711           & \cellcolor[HTML]{EFEFEF}\textbf{0.848}  & \multicolumn{1}{c|}{\cellcolor[HTML]{EFEFEF}\underline{0.738}}           & 0.835               & 0.677              \\
AdrenalMNIST          & 0.768           & \multicolumn{1}{c|}{0.435}          & 0.768             & \multicolumn{1}{c|}{0.435}             & 0.526           & 0.477           & \underline{0.815} & \multicolumn{1}{c|}{\textbf{0.715}} & 0.802          & 0.673           & \cellcolor[HTML]{EFEFEF}\textbf{0.819}  & \multicolumn{1}{c|}{\cellcolor[HTML]{EFEFEF}0.703}           & \underline{0.815}     & \underline{0.708}              \\
FractureMNIST         & 0.192           & \multicolumn{1}{c|}{0.107}          & 0.429             & \multicolumn{1}{c|}{0.201}             & 0.333           & 0.324          & \underline{0.517} & \multicolumn{1}{c|}{\textbf{0.381}} & 0.446          & 0.344           & \cellcolor[HTML]{EFEFEF}\textbf{0.521}  & \multicolumn{1}{c|}{\cellcolor[HTML]{EFEFEF}\underline{0.378}}           & 0.512               & 0.359              \\ \hline
\end{tabular}
}
\label{integration}

\end{table*}

\begin{figure*}[h]
    \centering
    \includegraphics[width=0.95\textwidth]{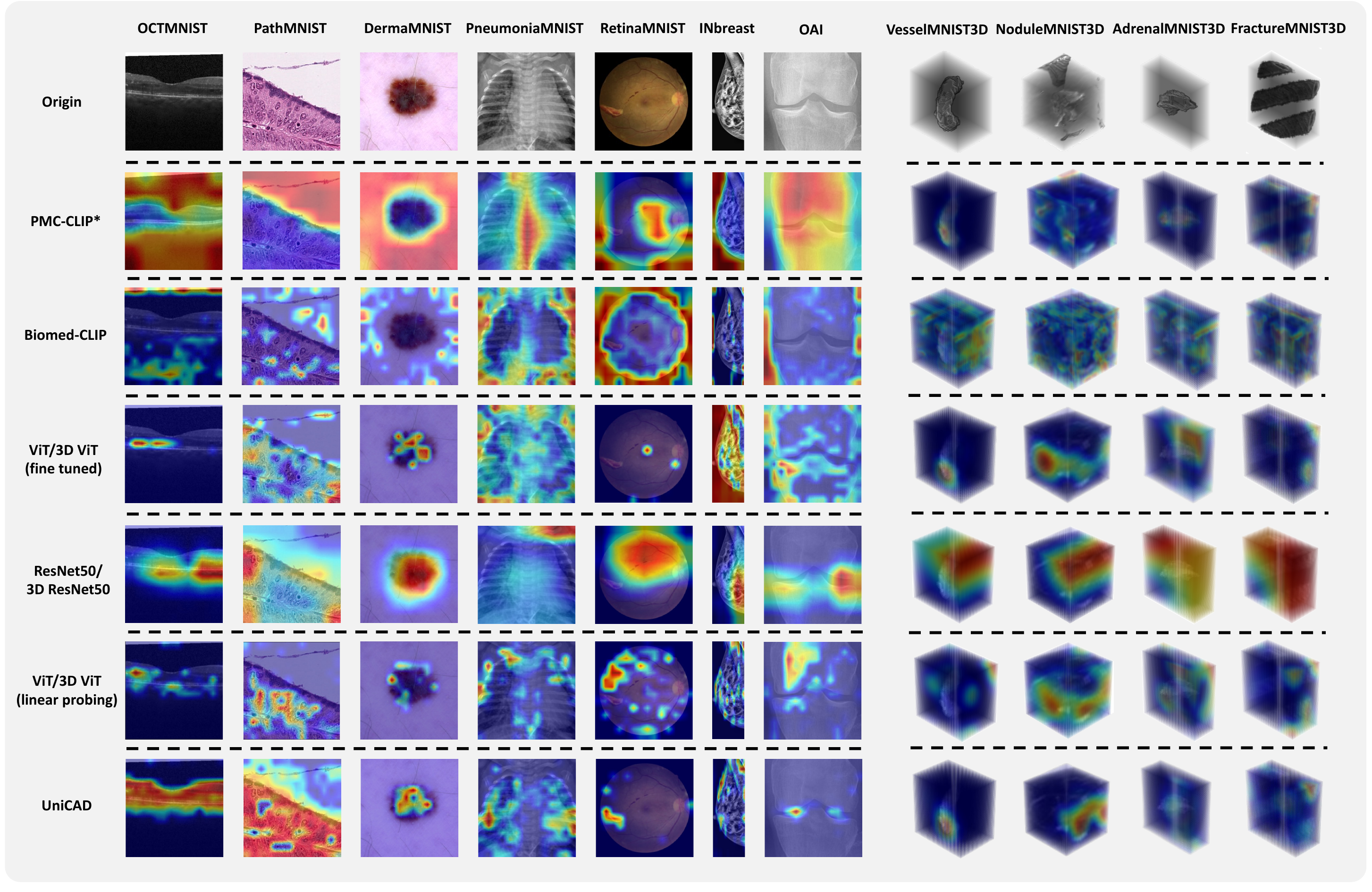}
    \caption{The attention visualization from different models for 2D and 3D medical images and diagnosis tasks.
    }
    \label{cam2d}
\end{figure*}

To visually interpret the diagnostic focus of different models, Figure \ref{cam2d} presents attention maps for selected approaches across our 11 single-label classification tasks.
The visualizations demonstrate that UniCAD exhibits more concentrated attention patterns on diagnostically relevant regions, contributing to its superior accuracy.
In contrast, the attention maps from standard ViT models appear more diffuse and less focused on clinically significant features.
Pre-trained general visual-language models such as PMC-CLIP and BiomedCLIP primarily attend to background elements and anatomical landmarks rather than pathological indicators, explaining their lower diagnostic performance.
For example, in the OAI dataset, UniCAD precisely highlights the joint spaces between bones where osteoarthritic changes manifest, while other models fail to target these critical diagnostic regions.
This focused attention on clinically relevant features demonstrates UniCAD's ability to effectively leverage the foundation model's visual representation capabilities for precise medical diagnosis.

\textbf{Multi-label Classification:}
To further validate our approach on more complex scenarios, we evaluate performance on multi-label classification using NIH Chest X-ray14, which requires predicting multiple co-occurring conditions with potential inter-label dependencies.
Table \ref{NIH} presents AUC-per-label results, where UniCAD and fully fine-tuned ViT consistently demonstrate strong performance across most thoracic abnormalities.
ResNet50 also achieves competitive results but generally trails behind UniCAD and ViT, further confirming the effectiveness of our approach in handling complex diagnostic scenarios.
Despite significant recent progress in general foundation models for medical imaging, these models still struggle to achieve strong performance on specialized diagnostic tasks, e.g., compared to UniCAD.

\begin{table*}[h]
\centering
\renewcommand{\arraystretch}{1.3}
\caption{Evaluation of multi-label classification performance, demonstrated on the NIH Chest X-ray14 dataset. The \colorbox{mygray}{gray} section represents our proposed method. \textbf{Bold} indicates the best result, and the \underline{underlined} number denotes the second-best result.}
\resizebox{\textwidth}{!}{
\begin{tabular}{l|ccccccc}
\hline
\multirow{3}{*}{Abnormality} & \multicolumn{7}{c}{AUC}                                                                                                                                                                                                                                             \\ \cline{2-8} 
                             & \multicolumn{3}{c|}{General Foundation Model}                                                                                                   & \multicolumn{2}{c|}{Task-Specific Model}                           & \multicolumn{2}{c}{Unified Multi-Task Model} \\
                             & \multicolumn{1}{c}{\begin{tabular}[c]{@{}c@{}}PMC-CLIP* \\ \citep{lin2023pmc}\end{tabular}} & \begin{tabular}[c]{@{}c@{}}BiomedCLIP \\ \citep{zhang2024biomedclip}\end{tabular} & \multicolumn{1}{c|}{\begin{tabular}[c]{@{}c@{}}Med-Flamingo \\ \citep{moor2023med}\end{tabular}} & \begin{tabular}[c]{@{}c@{}}ViT (fine-tuned) \\ \citep{dosovitskiy2020image}\end{tabular}& \multicolumn{1}{c|}{\begin{tabular}[c]{@{}c@{}}ResNet50 \\ \citep{he2016deep}\end{tabular}} & \cellcolor[HTML]{EFEFEF}UniCAD             & ViT (linear probing)   \\ \hline
Atelectasis                  & 0.528                                            & 0.452                                 & \multicolumn{1}{c|}{0.506}                           & \textbf{0.766}   & \multicolumn{1}{c|}{0.758}                      & \cellcolor[HTML]{EFEFEF}\underline{0.761}    & 0.676                   \\
Cardiomegaly                 & 0.530                                            & 0.620                                 & \multicolumn{1}{c|}{0.512}                           & \textbf{0.873}   & \multicolumn{1}{c|}{0.861}                      & \cellcolor[HTML]{EFEFEF}\underline{0.870}    & 0.711                   \\
Effusion                     & 0.472                                            & 0.556                                 & \multicolumn{1}{c|}{0.470}                           & \textbf{0.823}   & \multicolumn{1}{c|}{\underline{0.821}}            & \cellcolor[HTML]{EFEFEF}0.819              & 0.726                   \\
Infiltration                 & 0.495                                            & 0.538                                 & \multicolumn{1}{c|}{0.536}                           & \textbf{0.708}   & \multicolumn{1}{c|}{0.695}                      & \cellcolor[HTML]{EFEFEF}\underline{0.705}    & 0.672                   \\
Mass                         & 0.520                                            & 0.500                                 & \multicolumn{1}{c|}{0.552}                           & \textbf{0.796}   & \multicolumn{1}{c|}{0.786}                      & \cellcolor[HTML]{EFEFEF}\underline{0.794}    & 0.661                   \\
Nodule                       & 0.493                                            & 0.538                                 & \multicolumn{1}{c|}{0.468}                           & \textbf{0.750}   & \multicolumn{1}{c|}{\underline{0.742}}            & \cellcolor[HTML]{EFEFEF}0.737              & 0.663                   \\
Pneumonia                    & 0.496                                            & 0.500                                 & \multicolumn{1}{c|}{0.459}                           & \underline{0.704}  & \multicolumn{1}{c|}{0.664}                      & \cellcolor[HTML]{EFEFEF}\textbf{0.713}     & 0.623                   \\
Pneumothorax                 & 0.496                                            & 0.500                                 & \multicolumn{1}{c|}{0.531}                           & \textbf{0.855}   & \multicolumn{1}{c|}{0.837}                      & \cellcolor[HTML]{EFEFEF}\underline{0.849}    & 0.758                   \\
Consolidation                & 0.476                                            & 0.700                                 & \multicolumn{1}{c|}{0.464}                           & \textbf{0.741}   & \multicolumn{1}{c|}{0.721}                      & \cellcolor[HTML]{EFEFEF}\underline{0.727}    & 0.682                   \\
Edema                        & 0.500                                            & 0.508                                 & \multicolumn{1}{c|}{0.444}                           & \textbf{0.836}   & \multicolumn{1}{c|}{0.815}                      & \cellcolor[HTML]{EFEFEF}\underline{0.828}    & 0.775                   \\
Emphysema                    & 0.502                                            & 0.500                                 & \multicolumn{1}{c|}{0.420}                           & 0.870            & \multicolumn{1}{c|}{\textbf{0.890}}            & \cellcolor[HTML]{EFEFEF}\underline{0.884}    & 0.717                   \\
Fibrosis                     & 0.475                                            & 0.485                                 & \multicolumn{1}{c|}{0.530}                           & \underline{0.784}  & \multicolumn{1}{c|}{0.759}                      & \cellcolor[HTML]{EFEFEF}\textbf{0.801}     & 0.777                   \\
Pleural Thickening           & 0.491                                            & 0.500                                 & \multicolumn{1}{c|}{0.550}                           & \underline{0.752}  & \multicolumn{1}{c|}{0.742}                      & \cellcolor[HTML]{EFEFEF}\textbf{0.756}     & 0.709                   \\
Hernia                       & 0.474                                            & 0.492                                 & \multicolumn{1}{c|}{0.496}                           & 0.772            & \multicolumn{1}{c|}{0.827}                      & \cellcolor[HTML]{EFEFEF}\textbf{0.879}     & \underline{0.846}         \\ \hline
Average                      & 0.496                                            & 0.500                                 & \multicolumn{1}{c|}{0.496}                           & \underline{0.788}  & \multicolumn{1}{c|}{0.780}                      & \cellcolor[HTML]{EFEFEF}\textbf{0.795}     & 0.714                   \\ \hline
\end{tabular}
}
\label{NIH}
\end{table*}

\subsection{Accuracy and Memory Trade-offs}
\begin{figure*}[h]
    \centering
    \includegraphics[width=0.95\textwidth]{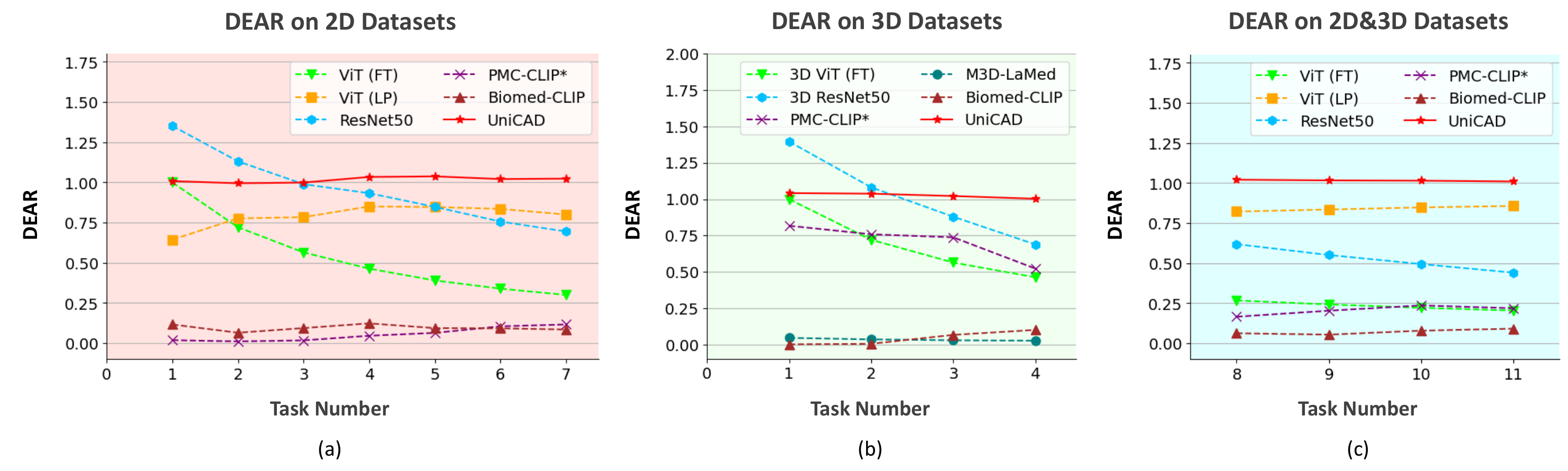}
    \caption{DEAR trends for multi-task CAD systems. 
    (a) Shows the change in DEAR for different methods as the number of task types increases on 2D datasets, where we evaluate up to 7 single-label classification tasks. 
    (b) Illustrates the change in DEAR for different methods as the number of task types increases on 3D datasets, with a maximum of 4 distinct tasks. 
    (c) Displays the change in DEAR for different methods when sequentially adding 3D tasks to a system that already supports all seven 2D tasks, demonstrating cross-dimensional scaling capabilities.
    Task indices in this figure are consistent with those presented in Table \ref{integration}.
    }
    \label{dear}
\end{figure*}

\begin{table}[]
\caption{Performance comparison of multi-task CAD systems across 2D and 3D tasks, including DEAR, average accuracy (Avg ACC), and GPU memory usage (GRAM). The \colorbox{mygray}{gray} section represents our proposed method. \textbf{Bold} indicates the best result, and the \underline{underlined} number denotes the second-best result.}
\resizebox{\linewidth}{!}{
\renewcommand{\arraystretch}{1.5}
\begin{tabular}{lccccc}
\hline
Method                               & Dim         & Task Number & DEAR              & Avg ACC           & GRAM              \\ \hline
\cellcolor[HTML]{EFEFEF}UniCAD                               & \cellcolor[HTML]{EFEFEF}2D$\backslash$3D & \cellcolor[HTML]{EFEFEF}11          & \cellcolor[HTML]{EFEFEF}\textbf{1.010}    & \cellcolor[HTML]{EFEFEF}\textbf{0.795}    & \cellcolor[HTML]{EFEFEF}\underline{0.93G} \\
ViT (fine-tuned) \citep{dosovitskiy2020image}    & 2D$\backslash$3D & 11          & 0.204             & \underline{0.788} & 4.48G             \\
ViT (linear probing)                  & 2D$\backslash$3D & 11          & \underline{0.857}             & 0.750             & \textbf{0.92G}    \\
ResNet50 \citep{he2016deep}            & 2D$\backslash$3D & 11          & 0.441       & 0.772             & 1.95G             \\ \hline
\cellcolor[HTML]{EFEFEF}UniCAD                               & \cellcolor[HTML]{EFEFEF}2D only              & \cellcolor[HTML]{EFEFEF}7           & \cellcolor[HTML]{EFEFEF}\textbf{1.023}    & \cellcolor[HTML]{EFEFEF}\textbf{0.808}    & \cellcolor[HTML]{EFEFEF}\textbf{0.92G}    \\
PMC-CLIP* \citep{lin2023pmc}           & 2D only               & 7           & \underline{0.116} & \underline{0.414} & \underline1.10G   \\
BiomedCLIP \citep{zhang2024biomedclip} & 2D only              & 7           & 0.085             & 0.373             & \underline1.10G   \\
Med-Flamingo \citep{moor2023med}       & 2D only              & 7           & 0.001             & 0.233             & 30.69G            \\ \hline
\cellcolor[HTML]{EFEFEF}UniCAD                               & \cellcolor[HTML]{EFEFEF}3D only               & \cellcolor[HTML]{EFEFEF}4           & \cellcolor[HTML]{EFEFEF}\textbf{1.003}    & \cellcolor[HTML]{EFEFEF}\textbf{0.774}    & \cellcolor[HTML]{EFEFEF}\textbf{0.92G}    \\
PMC-CLIP* \citep{lin2023pmc}           & 3D only              & 4           & \underline{0.523} & \underline{0.660}             & \underline{1.10G} \\
BiomedCLIP \citep{zhang2024biomedclip} & 3D only              & 4           & 0.101             & 0.382             & \underline{1.10G} \\
M3D-LaMed \citep{bai2024m3d}           & 3D only              & 4           & 0.027             & 0.563 & 13.02G             \\ \hline
\end{tabular}
}
\label{tab:DEAR}
\end{table}

The experiments discussed earlier compare the accuracy of different models across various medical image diagnosis tasks. 
However, for real-world clinical deployment, the hardware resource requirements of a CAD system must also be considered. 
To better evaluate the performance of multi-task CAD systems, we propose a new metric called Diagnostic Efficacy and Allocation Ratio (DEAR). 
For model $M$, the DEAR calculation across $N$ datasets is defined by the following equation:
\begin{equation}
    \text{DEAR}(M,N) = \left(\frac{\sum_{i=1}^{N} A_{M}^{i}}{\sum_{i=1}^{N} A_{\text{ViT}}^{i}}\right)^{k} \bigg/ \left(\frac{\sum_{i=1}^{N}E_{M}^{i}}{N\times E_{\text{ViT}}}\right).
\end{equation}
Here, $A_{M}^{i}$ represents the average accuracy of model $M$ on task $i$, and $E_{M}^{i}$ denotes the GPU memory consumption required to run model $M$ on the task. 
To account for task variation, we use a fully fine-tuned ViT model as the baseline for each $i$. 
Both accuracy ($A_{M}^{i}$) and memory consumption ($E_{M}^{i}$) are normalized by their respective baseline values, $A_{\text{ViT}}^{i}$ and $E_{\text{ViT}}$. 
The coefficient $k$ (=3) is empirically introduced to balance the trade-off between diagnostic efficacy and hardware resource requirements.

For the seven 2D and four 3D tasks used in Table \ref{integration}, we present the DEAR index curves for all methods as a function of the number of tasks in Figure \ref{dear}.
The number of tasks represents the maximum number of distinct diagnostic task types that can be simultaneously supported in a clinical workflow.
A higher number indicates that the system can concurrently handle more types of diagnosis tasks, while a lower number suggests the system can efficiently manage only a limited task range.
When this number equals 1, the system effectively functions as a single-task model.

As observed in the figure, UniCAD's DEAR value consistently remains around 1 across all task counts, demonstrating that it achieves comparable accuracy to individually fine-tuned ViT models while maintaining memory consumption similar to using a single ViT model.
This stability persists even as the number of tasks increases, highlighting UniCAD's scalability and efficiency advantage in multi-task clinical deployments.
As detailed in Table \ref{tab:DEAR}, UniCAD consistently leads in both average accuracy and GPU memory efficiency, highlighting the advantages of combining a shared general model with multiple lightweight task-specific experts.
In contrast, architectures employing multiple models for multiple tasks (e.g., ViT and ResNet50) maintain high accuracy but suffer from linearly increasing GPU memory usage as tasks accumulate, resulting in declining DEAR indices. 
Single-model approaches handling multiple tasks (e.g., CLIP variants and Med-Flamingo) demonstrate lower accuracy while often requiring substantial GPU memory, further compromising their deployment efficiency in clinical settings.

\subsection{Multi-Task Throughput in Clinical Workflows}
\label{exp4}

To evaluate the deployment and inference efficiency of our proposed UniCAD framework, we conduct a simulation experiment to mimic different task settings.
We randomly select 256 medical images and evenly distribute them according to the specified number of tasks.
For instance, when the task number is set to 1, all 256 images belong to the same task; when set to 8, each task receives 32 images.
We evaluate two processing modes: non-random flow and random flow.
In the non-random flow mode, the system processes all images from one task before moving to the next.
In the random flow mode, all 256 images are shuffled to simulate a mixed-task input stream.
This random flow mode more closely aligns with real-world clinical scenarios, where multiple tasks may arrive in a random order rather than being neatly batched by task type.

UniCAD supports random flow input, enabling simultaneous processing of multiple tasks within a single batch, thus allowing a larger batch size to accommodate images from different task types concurrently.
In contrast, the traditional approach to clinical practice (corresponding to the second category outlined in the Introduction) employs individually fine-tuned ViT models for each task, requiring all ViTs to be loaded onto the GPU.
This traditional approach can only increase the batch size for images from the same task, limiting its efficiency when handling heterogeneous datasets.
We exclude general multi-task models from this comparison due to their significantly lower diagnostic accuracy observed in our previous experiments, which makes them impractical for real-world clinical deployment.
For our deployment comparison, we use both ViT \textit{base} and ViT \textit{giga} as backbones to evaluate performance across different foundation model sizes, while in all other experiments throughout this paper, we use ViT \textit{base} as the default foundation model.

As shown in Table \ref{tab:computation}, UniCAD utilizes a larger batch size to accelerate processing as the task number increases.
This acceleration comes without significant increases in GPU memory usage due to the shared foundation model and the minimal additional parameters required for each task, making UniCAD more convenient for practical deployment.
Furthermore, the results show that when the vision foundation model evolves from ViT \textit{base} to ViT \textit{giga}, the performance gains remain consistent, demonstrating that UniCAD provides an efficient and scalable solution regardless of foundation model sizes.
In contrast, fine-tuned ViT models suffer from the lack of connection between different tasks, causing computational resources to increase dramatically as the number of tasks grows, which creates a significant bottleneck for real-world clinical deployment.

\begin{table}[t]
  \centering
  \caption{Performance is evaluated on 256 medical images, evenly distributed across tasks, using either a dataset-specific order or a randomized order to simulate clinical workflow. The \colorbox{mygray}{gray} section represents our proposed method.}

\resizebox{\linewidth}{!}{
\renewcommand{\arraystretch}{1.5}
\begin{tabular}{l|ccccccc}
\hline
                                                 &                                                                         &                                                                         &                                                                        & \multicolumn{2}{c}{base} & \multicolumn{2}{c}{giga} \\ \cline{5-8} 
\multirow{-2}{*}{Method}                         & \multirow{-2}{*}{\begin{tabular}[c]{@{}c@{}}Random\\ Flow\end{tabular}} & \multirow{-2}{*}{\begin{tabular}[c]{@{}c@{}}Task\\ Number\end{tabular}} & \multirow{-2}{*}{\begin{tabular}[c]{@{}c@{}}Batch\\ Size\end{tabular}} & Time        & GRAM       & Time        & GRAM       \\ \hline
                                                 &                                                                         & 1                                                                       & 1                                                                      & 1.79s       & 1.45G      & 22.51s      & 7.49G      \\
                                                 &                                                                         & 1                                                                       & 32                                                                     & 0.75s       & 1.73G      & 13.46s      & 8.43G      \\
                                                 &                                                                         & 2                                                                       & 1                                                                      & 1.70s       & 1.81G      & 22.49s      & 14.46G     \\
                                                 &                                                                         & 4                                                                       & 1                                                                      & 1.70s       & 2.51G      & 22.49s      & 28.37G     \\
\multirow{-5}{*}{ViT (fine-tuned)}               & \multirow{-5}{*}{\ding{56}}                                             & 8                                                                       & 1                                                                      & 1.70s       & 3.94G      & 22.47s      & 56.19G     \\ \hline
\rowcolor[HTML]{EFEFEF} 
\cellcolor[HTML]{EFEFEF}                         & \cellcolor[HTML]{EFEFEF}                                                & 1                                                                       & 1                                                                      & 2.97s       & 1.08G      & 24.94s      & 7.49G      \\
\rowcolor[HTML]{EFEFEF} 
\cellcolor[HTML]{EFEFEF}                         & \cellcolor[HTML]{EFEFEF}                                                & 1                                                                       & 32                                                                     & 0.90s       & 1.76G      & 14.99s      & 8.45G      \\
\rowcolor[HTML]{EFEFEF} 
\cellcolor[HTML]{EFEFEF}                         & \cellcolor[HTML]{EFEFEF}                                                & 2                                                                       & 32                                                                     & 0.90s       & 1.76G      & 14.90s      & 8.44G      \\
\rowcolor[HTML]{EFEFEF} 
\cellcolor[HTML]{EFEFEF}                         & \cellcolor[HTML]{EFEFEF}                                                & 4                                                                       & 32                                                                     & 0.91s       & 1.76G      & 14.91s      & 8.45G      \\
\rowcolor[HTML]{EFEFEF} 
\multirow{-5}{*}{\cellcolor[HTML]{EFEFEF}UniCAD} & \multirow{-5}{*}{\cellcolor[HTML]{EFEFEF}\ding{52}}                     & 8                                                                       & 32                                                                     & 0.90s       & 1.77G      & 14.91s      & 8.45G      \\ \hline
\end{tabular}
}
\label{tab:computation}%
\end{table}%

\subsection{Scaling Effects of Foundation Model Size}
\label{exp2}

\begin{figure*}[h]
    \centering
    \includegraphics[width=1\textwidth]{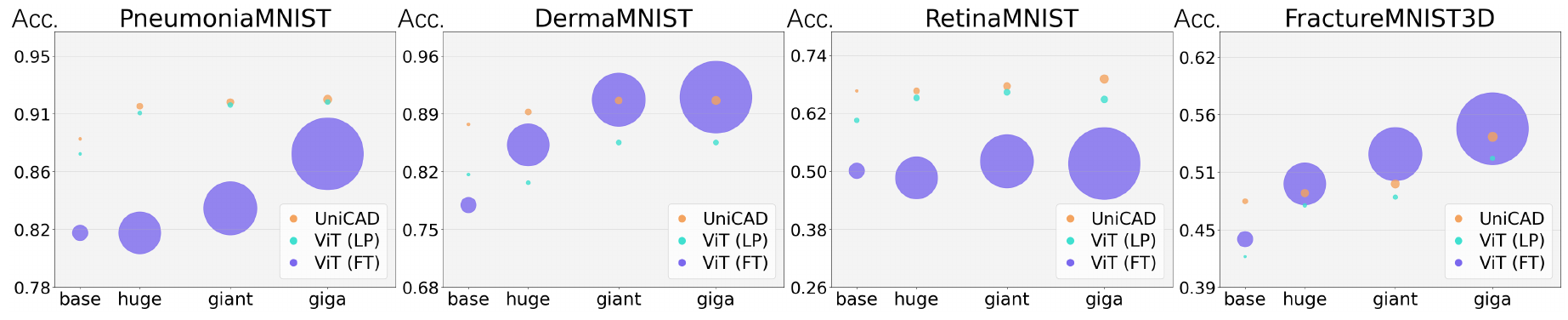}
    \caption{The accuracy trends of different methods as the parameter size of the foundation model increases. We utilize ViT \textit{base} (82M), ViT \textit{huge} (602M), ViT \textit{giant} (964M), and ViT \textit{giga} (1757M) to represent increasingly powerful vision foundation models pre-trained on natural images. The size of the circles in the figure corresponds to the number of trainable parameters. In the legend, "ViT (LP)" refers to the linear probing method, and "ViT (FT)" represents the fine-tuned ViT model.}

    \label{model_size}
\end{figure*}

To investigate the impact of stronger vision foundation models on UniCAD, we evaluate the performance of UniCAD across various model scales on four distinct datasets: PneumoniaMNIST, DermaMNIST, RetinaMNIST, and FractureMNIST3D. 
These datasets are selected because they exhibit significant changes in performance with increasing model parameters, whereas for other datasets our method has achieved very good results already, reaching a performance plateau with ViT \textit{base}.

As shown in Figure \ref{model_size}, the accuracy of UniCAD models consistently correlates with foundation model size.
As the ViT foundation model scales from \textit{base} to \textit{giga}, UniCAD demonstrates a steady performance improvement across all datasets.
This upward trajectory is particularly notable when compared to fine-tuned ViT and linear probing approaches, which frequently exhibit performance plateaus or even decrements as model size increases.
These comparative results indicate that UniCAD's architecture efficiently leverages the additional representational capacity of larger foundation models to enhance diagnostic accuracy while successfully avoiding overfitting.
The consistent scaling behavior suggests UniCAD can effectively harness future advances in vision foundation models with minimal architectural modifications.

Importantly, despite the increase in foundation model size, the number of trainable parameters within UniCAD remains relatively low.
While linear probing has fewer parameters overall, its performance is subpar and lacks stability across different model scales.
This parameter efficiency demonstrates that UniCAD's design achieves significant performance gains without a proportional increase in computational complexity.
In contrast, fine-tuned ViT requires a substantially larger number of trainable parameters as model size grows, resulting in significant increases in both training time and computational resource requirements.
These findings support the hypothesis that further scaling of pre-trained ViT models, when integrated into UniCAD, can deliver substantial performance improvements while requiring only minimal increases in trainable parameter count.

\section{Discussion and Conclusion}

We have proposed UniCAD, a unified architecture for multi-task computer-aided diagnosis that efficiently processes both 2D and 3D medical images.
By combining a shared ViT backbone with lightweight task-specific expert modules, UniCAD achieves high diagnostic accuracy across diverse imaging modalities with minimal computational overhead.
Unlike traditional approaches requiring separate large models for each task, our framework enables concurrent processing of multiple diagnostic tasks within a single system.
This approach facilitates rapid adaptation to new tasks without catastrophic forgetting while significantly reducing memory requirements.
Furthermore, UniCAD promotes collaborative development through its shareable expert modules.
These lightweight modules enable institutions to exchange diagnostic capabilities without sharing patient data, creating a secure ecosystem for collaborative advancement of medical AI technology across institutions.

Despite these advantages, several future directions could further enhance UniCAD's applicability, scalability, and generalization.
A particularly promising direction is the integration of multi-modal data, including electronic health records (EHRs), clinical notes, and radiology reports alongside medical images.
Currently, our approach utilizes ViTs rather than large language models (LLMs) because ViTs excel at capturing the spatial features essential for accurate medical image diagnosis.
However, this design choice means the current ViT backbone in UniCAD cannot directly process textual information from clinical documents.
To address this limitation, future versions could incorporate a dedicated text encoder that works in parallel with the visual foundation model.
These parallel encoders could then be aligned through cross-modal attention mechanisms or CLIP-style contrastive objectives to create a unified representation space.
This architecture would enable the system to jointly reason over both visual findings and clinical history, significantly enhancing decision-making capabilities in complex clinical scenarios where image data alone may be insufficient.

Despite UniCAD's parameter efficiency through LoRA, a significant challenge remains in the data requirements for training high-quality expert modules.
While the parameter updates are minimal, developing effective experts still necessitates curated and well-annotated datasets—a particular bottleneck for rare conditions or underrepresented tasks in medical imaging.
To address this limitation, future research should explore meta-learning and few-shot adaptation strategies that could significantly reduce the annotation burden for new diagnostic tasks.
These approaches could enable rapid adaptation to specialized diagnostic requirements with limited labeled examples, making the system more accessible for rare disease detection.
Complementary to these efforts, self-supervised learning techniques could further enhance model robustness by leveraging large quantities of unlabeled medical images, creating more generalizable representations for downstream fine-tuning.
Together, these advancements would strengthen UniCAD's applicability in resource-constrained settings and expand its utility across a broader spectrum of clinical applications, including those with limited labeled data availability.

\printcredits

\bibliographystyle{cas-model2-names}

\bibliography{cas-refs}

\begin{thebibliography}{36}
\expandafter\ifx\csname natexlab\endcsname\relax\def\natexlab#1{#1}\fi
\providecommand{\url}[1]{\texttt{#1}}
\providecommand{\href}[2]{#2}
\providecommand{\path}[1]{#1}
\providecommand{\DOIprefix}{doi:}
\providecommand{\ArXivprefix}{arXiv:}
\providecommand{\URLprefix}{URL: }
\providecommand{\Pubmedprefix}{pmid:}
\providecommand{\doi}[1]{\href{http://dx.doi.org/#1}{\path{#1}}}
\providecommand{\Pubmed}[1]{\href{pmid:#1}{\path{#1}}}
\providecommand{\bibinfo}[2]{#2}
\ifx\xfnm\relax \def\xfnm[#1]{\unskip,\space#1}\fi
\bibitem[{Bai et~al.(2024)Bai, Du, Huang, Meng and Zhao}]{bai2024m3d}
\bibinfo{author}{Bai, F.}, \bibinfo{author}{Du, Y.}, \bibinfo{author}{Huang, T.}, \bibinfo{author}{Meng, M.Q.H.}, \bibinfo{author}{Zhao, B.}, \bibinfo{year}{2024}.
\newblock \bibinfo{title}{M3d: Advancing 3d medical image analysis with multi-modal large language models}.
\newblock \bibinfo{journal}{arXiv preprint arXiv:2404.00578} .
\bibitem[{Dehghani et~al.(2023)Dehghani, Djolonga, Mustafa, Padlewski, Heek, Gilmer, Steiner, Caron, Geirhos, Alabdulmohsin et~al.}]{dehghani2023scaling}
\bibinfo{author}{Dehghani, M.}, \bibinfo{author}{Djolonga, J.}, \bibinfo{author}{Mustafa, B.}, \bibinfo{author}{Padlewski, P.}, \bibinfo{author}{Heek, J.}, \bibinfo{author}{Gilmer, J.}, \bibinfo{author}{Steiner, A.P.}, \bibinfo{author}{Caron, M.}, \bibinfo{author}{Geirhos, R.}, \bibinfo{author}{Alabdulmohsin, I.}, et~al., \bibinfo{year}{2023}.
\newblock \bibinfo{title}{Scaling vision transformers to 22 billion parameters}, in: \bibinfo{booktitle}{International Conference on Machine Learning}, \bibinfo{organization}{PMLR}. pp. \bibinfo{pages}{7480--7512}.
\bibitem[{Dosovitskiy et~al.(2020)Dosovitskiy, Beyer, Kolesnikov, Weissenborn, Zhai, Unterthiner, Dehghani, Minderer, Heigold, Gelly et~al.}]{dosovitskiy2020image}
\bibinfo{author}{Dosovitskiy, A.}, \bibinfo{author}{Beyer, L.}, \bibinfo{author}{Kolesnikov, A.}, \bibinfo{author}{Weissenborn, D.}, \bibinfo{author}{Zhai, X.}, \bibinfo{author}{Unterthiner, T.}, \bibinfo{author}{Dehghani, M.}, \bibinfo{author}{Minderer, M.}, \bibinfo{author}{Heigold, G.}, \bibinfo{author}{Gelly, S.}, et~al., \bibinfo{year}{2020}.
\newblock \bibinfo{title}{An image is worth 16x16 words: Transformers for image recognition at scale}, in: \bibinfo{booktitle}{International Conference on Learning Representations}.
\bibitem[{He et~al.(2016)He, Zhang, Ren and Sun}]{he2016deep}
\bibinfo{author}{He, K.}, \bibinfo{author}{Zhang, X.}, \bibinfo{author}{Ren, S.}, \bibinfo{author}{Sun, J.}, \bibinfo{year}{2016}.
\newblock \bibinfo{title}{Deep residual learning for image recognition}, in: \bibinfo{booktitle}{Proceedings of the IEEE conference on computer vision and pattern recognition}, pp. \bibinfo{pages}{770--778}.
\bibitem[{Heker and Greenspan(2020)}]{heker2020joint}
\bibinfo{author}{Heker, M.}, \bibinfo{author}{Greenspan, H.}, \bibinfo{year}{2020}.
\newblock \bibinfo{title}{Joint liver lesion segmentation and classification via transfer learning}, in: \bibinfo{booktitle}{Medical Imaging with Deep Learning}.
\bibitem[{Hu et~al.()Hu, Wallis, Allen-Zhu, Li, Wang, Wang, Chen et~al.}]{hulora}
\bibinfo{author}{Hu, E.J.}, \bibinfo{author}{Wallis, P.}, \bibinfo{author}{Allen-Zhu, Z.}, \bibinfo{author}{Li, Y.}, \bibinfo{author}{Wang, S.}, \bibinfo{author}{Wang, L.}, \bibinfo{author}{Chen, W.}, et~al., .
\newblock \bibinfo{title}{Lora: Low-rank adaptation of large language models}, in: \bibinfo{booktitle}{International Conference on Learning Representations}.
\bibitem[{Kellgren and Lawrence(1957)}]{kellgren1957radiological}
\bibinfo{author}{Kellgren, J.}, \bibinfo{author}{Lawrence, J.}, \bibinfo{year}{1957}.
\newblock \bibinfo{title}{Radiological assessment of osteo-arthrosis}.
\newblock \bibinfo{journal}{Annals of the rheumatic diseases} \bibinfo{volume}{16}, \bibinfo{pages}{494}.
\bibitem[{Kirillov et~al.(2023)Kirillov, Mintun, Ravi, Mao, Rolland, Gustafson, Xiao, Whitehead, Berg, Lo et~al.}]{kirillov2023segment}
\bibinfo{author}{Kirillov, A.}, \bibinfo{author}{Mintun, E.}, \bibinfo{author}{Ravi, N.}, \bibinfo{author}{Mao, H.}, \bibinfo{author}{Rolland, C.}, \bibinfo{author}{Gustafson, L.}, \bibinfo{author}{Xiao, T.}, \bibinfo{author}{Whitehead, S.}, \bibinfo{author}{Berg, A.C.}, \bibinfo{author}{Lo, W.Y.}, et~al., \bibinfo{year}{2023}.
\newblock \bibinfo{title}{Segment anything}, in: \bibinfo{booktitle}{Proceedings of the IEEE/CVF international conference on computer vision}, pp. \bibinfo{pages}{4015--4026}.
\bibitem[{Koleilat et~al.(2024)Koleilat, Asgariandehkordi, Rivaz and Xiao}]{koleilat2024medclip}
\bibinfo{author}{Koleilat, T.}, \bibinfo{author}{Asgariandehkordi, H.}, \bibinfo{author}{Rivaz, H.}, \bibinfo{author}{Xiao, Y.}, \bibinfo{year}{2024}.
\newblock \bibinfo{title}{Medclip-sam: Bridging text and image towards universal medical image segmentation}, in: \bibinfo{booktitle}{International Conference on Medical Image Computing and Computer-Assisted Intervention}, \bibinfo{organization}{Springer}. pp. \bibinfo{pages}{643--653}.
\bibitem[{Lai et~al.(2024)Lai, Wu, Chen, Zhou and Hovakimyan}]{lai2024residual}
\bibinfo{author}{Lai, Z.}, \bibinfo{author}{Wu, J.}, \bibinfo{author}{Chen, S.}, \bibinfo{author}{Zhou, Y.}, \bibinfo{author}{Hovakimyan, N.}, \bibinfo{year}{2024}.
\newblock \bibinfo{title}{Residual-based language models are free boosters for biomedical imaging}.
\newblock \bibinfo{journal}{CoRR} .
\bibitem[{Li et~al.(2023a)Li, Wong, Zhang, Usuyama, Liu, Yang, Naumann, Poon and Gao}]{li2023llava}
\bibinfo{author}{Li, C.}, \bibinfo{author}{Wong, C.}, \bibinfo{author}{Zhang, S.}, \bibinfo{author}{Usuyama, N.}, \bibinfo{author}{Liu, H.}, \bibinfo{author}{Yang, J.}, \bibinfo{author}{Naumann, T.}, \bibinfo{author}{Poon, H.}, \bibinfo{author}{Gao, J.}, \bibinfo{year}{2023}a.
\newblock \bibinfo{title}{Llava-med: Training a large language-and-vision assistant for biomedicine in one day}.
\newblock \bibinfo{journal}{Advances in Neural Information Processing Systems} \bibinfo{volume}{36}, \bibinfo{pages}{28541--28564}.
\bibitem[{Li et~al.(2023b)Li, Li, Li, Wang, Guo, Lu, Jin, Zhang and Hong}]{li2023lvit}
\bibinfo{author}{Li, Z.}, \bibinfo{author}{Li, Y.}, \bibinfo{author}{Li, Q.}, \bibinfo{author}{Wang, P.}, \bibinfo{author}{Guo, D.}, \bibinfo{author}{Lu, L.}, \bibinfo{author}{Jin, D.}, \bibinfo{author}{Zhang, Y.}, \bibinfo{author}{Hong, Q.}, \bibinfo{year}{2023}b.
\newblock \bibinfo{title}{Lvit: language meets vision transformer in medical image segmentation}.
\newblock \bibinfo{journal}{IEEE transactions on medical imaging} \bibinfo{volume}{43}, \bibinfo{pages}{96--107}.
\bibitem[{Liberman and Menell(2002)}]{LauraLiberman2002BreastIR}
\bibinfo{author}{Liberman, L.}, \bibinfo{author}{Menell, J.H.}, \bibinfo{year}{2002}.
\newblock \bibinfo{title}{Breast imaging reporting and data system (bi-rads).}
\newblock \bibinfo{journal}{Radiologic Clinics of North America} \bibinfo{volume}{40}, \bibinfo{pages}{409--430}.
\bibitem[{Lin et~al.(2023)Lin, Zhao, Zhang, Wu, Zhang, Wang and Xie}]{lin2023pmc}
\bibinfo{author}{Lin, W.}, \bibinfo{author}{Zhao, Z.}, \bibinfo{author}{Zhang, X.}, \bibinfo{author}{Wu, C.}, \bibinfo{author}{Zhang, Y.}, \bibinfo{author}{Wang, Y.}, \bibinfo{author}{Xie, W.}, \bibinfo{year}{2023}.
\newblock \bibinfo{title}{Pmc-clip: Contrastive language-image pre-training using biomedical documents}, in: \bibinfo{booktitle}{International Conference on Medical Image Computing and Computer-Assisted Intervention}, \bibinfo{organization}{Springer}. pp. \bibinfo{pages}{525--536}.
\bibitem[{Liu et~al.(2023)Liu, Wu, Zhao, Zhu, Xu, Tian and Zheng}]{liu2023moelora}
\bibinfo{author}{Liu, Q.}, \bibinfo{author}{Wu, X.}, \bibinfo{author}{Zhao, X.}, \bibinfo{author}{Zhu, Y.}, \bibinfo{author}{Xu, D.}, \bibinfo{author}{Tian, F.}, \bibinfo{author}{Zheng, Y.}, \bibinfo{year}{2023}.
\newblock \bibinfo{title}{Moelora: An moe-based parameter efficient fine-tuning method for multi-task medical applications}.
\newblock \bibinfo{journal}{CoRR} .
\bibitem[{Liu et~al.(2020)Liu, Jain, Eng, Way, Lee, Bui, Kanada, de~Oliveira~Marinho, Gallegos, Gabriele et~al.}]{liu2020deep}
\bibinfo{author}{Liu, Y.}, \bibinfo{author}{Jain, A.}, \bibinfo{author}{Eng, C.}, \bibinfo{author}{Way, D.H.}, \bibinfo{author}{Lee, K.}, \bibinfo{author}{Bui, P.}, \bibinfo{author}{Kanada, K.}, \bibinfo{author}{de~Oliveira~Marinho, G.}, \bibinfo{author}{Gallegos, J.}, \bibinfo{author}{Gabriele, S.}, et~al., \bibinfo{year}{2020}.
\newblock \bibinfo{title}{A deep learning system for differential diagnosis of skin diseases}.
\newblock \bibinfo{journal}{Nature medicine} \bibinfo{volume}{26}, \bibinfo{pages}{900--908}.
\bibitem[{Ma et~al.(2024)Ma, He, Li, Han, You and Wang}]{ma2024segment}
\bibinfo{author}{Ma, J.}, \bibinfo{author}{He, Y.}, \bibinfo{author}{Li, F.}, \bibinfo{author}{Han, L.}, \bibinfo{author}{You, C.}, \bibinfo{author}{Wang, B.}, \bibinfo{year}{2024}.
\newblock \bibinfo{title}{Segment anything in medical images}.
\newblock \bibinfo{journal}{Nature Communications} \bibinfo{volume}{15}, \bibinfo{pages}{654}.
\bibitem[{McKinney et~al.(2020)McKinney, Sieniek, Godbole, Godwin, Antropova, Ashrafian, Back, Chesus, Corrado, Darzi et~al.}]{mckinney2020international}
\bibinfo{author}{McKinney, S.M.}, \bibinfo{author}{Sieniek, M.}, \bibinfo{author}{Godbole, V.}, \bibinfo{author}{Godwin, J.}, \bibinfo{author}{Antropova, N.}, \bibinfo{author}{Ashrafian, H.}, \bibinfo{author}{Back, T.}, \bibinfo{author}{Chesus, M.}, \bibinfo{author}{Corrado, G.S.}, \bibinfo{author}{Darzi, A.}, et~al., \bibinfo{year}{2020}.
\newblock \bibinfo{title}{International evaluation of an ai system for breast cancer screening}.
\newblock \bibinfo{journal}{Nature} \bibinfo{volume}{577}, \bibinfo{pages}{89--94}.
\bibitem[{Miao et~al.(2024)Miao, Chen, Zhang, Chuai, Li and Heng}]{miao2024cross}
\bibinfo{author}{Miao, J.}, \bibinfo{author}{Chen, C.}, \bibinfo{author}{Zhang, K.}, \bibinfo{author}{Chuai, J.}, \bibinfo{author}{Li, Q.}, \bibinfo{author}{Heng, P.A.}, \bibinfo{year}{2024}.
\newblock \bibinfo{title}{Cross prompting consistency with segment anything model for semi-supervised medical image segmentation}, in: \bibinfo{booktitle}{International Conference on Medical Image Computing and Computer-Assisted Intervention}, \bibinfo{organization}{Springer}. pp. \bibinfo{pages}{167--177}.
\bibitem[{Moor et~al.(2023)Moor, Huang, Wu, Yasunaga, Dalmia, Leskovec, Zakka, Reis and Rajpurkar}]{moor2023med}
\bibinfo{author}{Moor, M.}, \bibinfo{author}{Huang, Q.}, \bibinfo{author}{Wu, S.}, \bibinfo{author}{Yasunaga, M.}, \bibinfo{author}{Dalmia, Y.}, \bibinfo{author}{Leskovec, J.}, \bibinfo{author}{Zakka, C.}, \bibinfo{author}{Reis, E.P.}, \bibinfo{author}{Rajpurkar, P.}, \bibinfo{year}{2023}.
\newblock \bibinfo{title}{Med-flamingo: a multimodal medical few-shot learner}, in: \bibinfo{booktitle}{Machine Learning for Health (ML4H)}, \bibinfo{organization}{PMLR}. pp. \bibinfo{pages}{353--367}.
\bibitem[{Moreira et~al.(2012)Moreira, Amaral, Domingues, Cardoso, Cardoso and Cardoso}]{InsMoreira2012INbreastTA}
\bibinfo{author}{Moreira, I.}, \bibinfo{author}{Amaral, I.}, \bibinfo{author}{Domingues, I.}, \bibinfo{author}{Cardoso, A.}, \bibinfo{author}{Cardoso, M.J.}, \bibinfo{author}{Cardoso, J.S.}, \bibinfo{year}{2012}.
\newblock \bibinfo{title}{Inbreast: toward a full-field digital mammographic database.}
\newblock \bibinfo{journal}{Academic Radiology} \bibinfo{volume}{19}, \bibinfo{pages}{236--248}.
\bibitem[{Radford et~al.(2021)Radford, Kim, Hallacy, Ramesh, Goh, Agarwal, Sastry, Askell, Mishkin, Clark et~al.}]{radford2021learning}
\bibinfo{author}{Radford, A.}, \bibinfo{author}{Kim, J.W.}, \bibinfo{author}{Hallacy, C.}, \bibinfo{author}{Ramesh, A.}, \bibinfo{author}{Goh, G.}, \bibinfo{author}{Agarwal, S.}, \bibinfo{author}{Sastry, G.}, \bibinfo{author}{Askell, A.}, \bibinfo{author}{Mishkin, P.}, \bibinfo{author}{Clark, J.}, et~al., \bibinfo{year}{2021}.
\newblock \bibinfo{title}{Learning transferable visual models from natural language supervision}, in: \bibinfo{booktitle}{International Conference on Machine Learning}, \bibinfo{organization}{PMLR}. pp. \bibinfo{pages}{8748--8763}.
\bibitem[{Razzak et~al.(2018)Razzak, Naz and Zaib}]{razzak2018deep}
\bibinfo{author}{Razzak, M.I.}, \bibinfo{author}{Naz, S.}, \bibinfo{author}{Zaib, A.}, \bibinfo{year}{2018}.
\newblock \bibinfo{title}{Deep learning for medical image processing: Overview, challenges and the future}.
\newblock \bibinfo{journal}{Classification in BioApps: Automation of Decision Making} , \bibinfo{pages}{323--350}.
\bibitem[{Shaharabany et~al.(2023)Shaharabany, Dahan, Giryes and Wolf}]{shaharabany2023autosam}
\bibinfo{author}{Shaharabany, T.}, \bibinfo{author}{Dahan, A.}, \bibinfo{author}{Giryes, R.}, \bibinfo{author}{Wolf, L.}, \bibinfo{year}{2023}.
\newblock \bibinfo{title}{Autosam: Adapting sam to medical images by overloading the prompt encoder}.
\newblock \bibinfo{journal}{arXiv preprint arXiv:2306.06370} .
\bibitem[{Shen et~al.(2023)Shen, Ouyang, Xiao, Cheng, Shen and Wang}]{shen2023image}
\bibinfo{author}{Shen, Z.}, \bibinfo{author}{Ouyang, X.}, \bibinfo{author}{Xiao, B.}, \bibinfo{author}{Cheng, J.Z.}, \bibinfo{author}{Shen, D.}, \bibinfo{author}{Wang, Q.}, \bibinfo{year}{2023}.
\newblock \bibinfo{title}{Image synthesis with disentangled attributes for chest x-ray nodule augmentation and detection}.
\newblock \bibinfo{journal}{Medical image analysis} \bibinfo{volume}{84}, \bibinfo{pages}{102708}.
\bibitem[{Shi et~al.(2024)Shi, Kim, Li, Li and Pfister}]{shi2024mora}
\bibinfo{author}{Shi, Z.}, \bibinfo{author}{Kim, J.}, \bibinfo{author}{Li, W.}, \bibinfo{author}{Li, Y.}, \bibinfo{author}{Pfister, H.}, \bibinfo{year}{2024}.
\newblock \bibinfo{title}{Mora: Lora guided multi-modal disease diagnosis with missing modality}, in: \bibinfo{booktitle}{International Conference on Medical Image Computing and Computer-Assisted Intervention}, \bibinfo{organization}{Springer}. pp. \bibinfo{pages}{273--282}.
\bibitem[{Wang et~al.(2023)Wang, Guo, Ye, Deng, Cheng, Li, Chen, Su, Huang, Shen et~al.}]{wang2023sam}
\bibinfo{author}{Wang, H.}, \bibinfo{author}{Guo, S.}, \bibinfo{author}{Ye, J.}, \bibinfo{author}{Deng, Z.}, \bibinfo{author}{Cheng, J.}, \bibinfo{author}{Li, T.}, \bibinfo{author}{Chen, J.}, \bibinfo{author}{Su, Y.}, \bibinfo{author}{Huang, Z.}, \bibinfo{author}{Shen, Y.}, et~al., \bibinfo{year}{2023}.
\newblock \bibinfo{title}{Sam-med3d: towards general-purpose segmentation models for volumetric medical images}.
\newblock \bibinfo{journal}{arXiv preprint arXiv:2310.15161} .
\bibitem[{Wang et~al.(2017)Wang, Peng, Lu, Lu, Bagheri and Summers}]{wang2017chestx}
\bibinfo{author}{Wang, X.}, \bibinfo{author}{Peng, Y.}, \bibinfo{author}{Lu, L.}, \bibinfo{author}{Lu, Z.}, \bibinfo{author}{Bagheri, M.}, \bibinfo{author}{Summers, R.M.}, \bibinfo{year}{2017}.
\newblock \bibinfo{title}{Chestx-ray8: Hospital-scale chest x-ray database and benchmarks on weakly-supervised classification and localization of common thorax diseases}, in: \bibinfo{booktitle}{Proceedings of the IEEE conference on computer vision and pattern recognition}, pp. \bibinfo{pages}{2097--2106}.
\bibitem[{Wang et~al.(2022)Wang, Wu, Agarwal and Sun}]{wang2022medclip}
\bibinfo{author}{Wang, Z.}, \bibinfo{author}{Wu, Z.}, \bibinfo{author}{Agarwal, D.}, \bibinfo{author}{Sun, J.}, \bibinfo{year}{2022}.
\newblock \bibinfo{title}{Medclip: Contrastive learning from unpaired medical images and text}, in: \bibinfo{booktitle}{Proceedings of the Conference on Empirical Methods in Natural Language Processing. Conference on Empirical Methods in Natural Language Processing}, p. \bibinfo{pages}{3876}.
\bibitem[{Wightman(2019)}]{rw2019timm}
\bibinfo{author}{Wightman, R.}, \bibinfo{year}{2019}.
\newblock \bibinfo{title}{Pytorch image models}.
\newblock \bibinfo{howpublished}{\url{https://github.com/rwightman/pytorch-image-models}}.
\newblock \DOIprefix\doi{10.5281/zenodo.4414861}.
\bibitem[{Wu et~al.(2023)Wu, Zhang, Zhang, Wang and Xie}]{wu2023medklip}
\bibinfo{author}{Wu, C.}, \bibinfo{author}{Zhang, X.}, \bibinfo{author}{Zhang, Y.}, \bibinfo{author}{Wang, Y.}, \bibinfo{author}{Xie, W.}, \bibinfo{year}{2023}.
\newblock \bibinfo{title}{Medklip: Medical knowledge enhanced language-image pre-training}.
\newblock \bibinfo{journal}{Proceedings of the IEEE/CVF International Conference on Computer Vision} .
\bibitem[{Yang et~al.(2023)Yang, Shi, Wei, Liu, Zhao, Ke, Pfister and Ni}]{medmnistv2}
\bibinfo{author}{Yang, J.}, \bibinfo{author}{Shi, R.}, \bibinfo{author}{Wei, D.}, \bibinfo{author}{Liu, Z.}, \bibinfo{author}{Zhao, L.}, \bibinfo{author}{Ke, B.}, \bibinfo{author}{Pfister, H.}, \bibinfo{author}{Ni, B.}, \bibinfo{year}{2023}.
\newblock \bibinfo{title}{Medmnist v2-a large-scale lightweight benchmark for 2d and 3d biomedical image classification}.
\newblock \bibinfo{journal}{Scientific Data} \bibinfo{volume}{10}, \bibinfo{pages}{41}.
\bibitem[{Zhang and Liu(2023)}]{zhang2023customized}
\bibinfo{author}{Zhang, K.}, \bibinfo{author}{Liu, D.}, \bibinfo{year}{2023}.
\newblock \bibinfo{title}{Customized segment anything model for medical image segmentation}.
\newblock \bibinfo{journal}{arXiv preprint arXiv:2304.13785} .
\bibitem[{Zhang et~al.(2024)Zhang, Xu, Usuyama, Xu, Bagga, Tinn, Preston, Rao, Wei, Valluri, Wong, Tupini, Wang, Mazzola, Shukla, Liden, Gao, Crabtree, Piening, Bifulco, Lungren, Naumann, Wang and Poon}]{zhang2024biomedclip}
\bibinfo{author}{Zhang, S.}, \bibinfo{author}{Xu, Y.}, \bibinfo{author}{Usuyama, N.}, \bibinfo{author}{Xu, H.}, \bibinfo{author}{Bagga, J.}, \bibinfo{author}{Tinn, R.}, \bibinfo{author}{Preston, S.}, \bibinfo{author}{Rao, R.}, \bibinfo{author}{Wei, M.}, \bibinfo{author}{Valluri, N.}, \bibinfo{author}{Wong, C.}, \bibinfo{author}{Tupini, A.}, \bibinfo{author}{Wang, Y.}, \bibinfo{author}{Mazzola, M.}, \bibinfo{author}{Shukla, S.}, \bibinfo{author}{Liden, L.}, \bibinfo{author}{Gao, J.}, \bibinfo{author}{Crabtree, A.}, \bibinfo{author}{Piening, B.}, \bibinfo{author}{Bifulco, C.}, \bibinfo{author}{Lungren, M.P.}, \bibinfo{author}{Naumann, T.}, \bibinfo{author}{Wang, S.}, \bibinfo{author}{Poon, H.}, \bibinfo{year}{2024}.
\newblock \bibinfo{title}{A multimodal biomedical foundation model trained from fifteen million image–text pairs}.
\newblock \bibinfo{journal}{NEJM AI} \bibinfo{volume}{2}.
\newblock \URLprefix \url{https://ai.nejm.org/doi/full/10.1056/AIoa2400640}, \DOIprefix\doi{10.1056/AIoa2400640}.
\bibitem[{Zhou et~al.(2024)Zhou, Du, Li, Yao, Zhang and Wang}]{zhou2024reprogramming}
\bibinfo{author}{Zhou, Y.}, \bibinfo{author}{Du, S.}, \bibinfo{author}{Li, H.}, \bibinfo{author}{Yao, J.}, \bibinfo{author}{Zhang, Y.}, \bibinfo{author}{Wang, Y.}, \bibinfo{year}{2024}.
\newblock \bibinfo{title}{Reprogramming distillation for medical foundation models}, in: \bibinfo{booktitle}{International Conference on Medical Image Computing and Computer-Assisted Intervention}, \bibinfo{organization}{Springer}. pp. \bibinfo{pages}{533--543}.
\bibitem[{Zhu et~al.(2024)Zhu, Shen, Zhao, Wang, Wang, Zhao, Shen and Wang}]{zhu2024melo}
\bibinfo{author}{Zhu, Y.}, \bibinfo{author}{Shen, Z.}, \bibinfo{author}{Zhao, Z.}, \bibinfo{author}{Wang, S.}, \bibinfo{author}{Wang, X.}, \bibinfo{author}{Zhao, X.}, \bibinfo{author}{Shen, D.}, \bibinfo{author}{Wang, Q.}, \bibinfo{year}{2024}.
\newblock \bibinfo{title}{Melo: Low-rank adaptation is better than fine-tuning for medical image diagnosis}, in: \bibinfo{booktitle}{2024 IEEE International Symposium on Biomedical Imaging (ISBI)}, \bibinfo{organization}{IEEE}. pp. \bibinfo{pages}{1--5}.

\end{thebibliography}



\end{document}